\documentclass{article}

\usepackage{PRIMEarxiv}
\usepackage{natbib}
\usepackage[utf8]{inputenc}
\usepackage[T1]{fontenc}
\usepackage{hyperref}
\usepackage{url}
\usepackage{booktabs}
\usepackage{amsfonts}
\usepackage{amssymb}
\usepackage{nicefrac}
\usepackage{microtype}
\usepackage{graphicx}
\usepackage{multirow}
\usepackage{array}
\usepackage{xcolor}
\usepackage{listings}
\usepackage{tikz}
\usetikzlibrary{arrows.meta,positioning,calc,fit,backgrounds}
\usepackage{placeins}

\newsavebox{\fitbox}
\newcommand{\fitwidth}[1]{%
  \sbox{\fitbox}{#1}%
  \ifdim\wd\fitbox>\textwidth
    \resizebox{\textwidth}{!}{\usebox{\fitbox}}%
  \else
    \usebox{\fitbox}%
  \fi
}

\newcommand{\rcpt}[1]{#1}

\definecolor{codegreen}{rgb}{0,0.6,0}
\definecolor{codegray}{rgb}{0.5,0.5,0.5}
\definecolor{codepurple}{rgb}{0.58,0,0.82}
\definecolor{backcolour}{rgb}{0.97,0.97,0.97}

\title{\includegraphics[width=\linewidth]{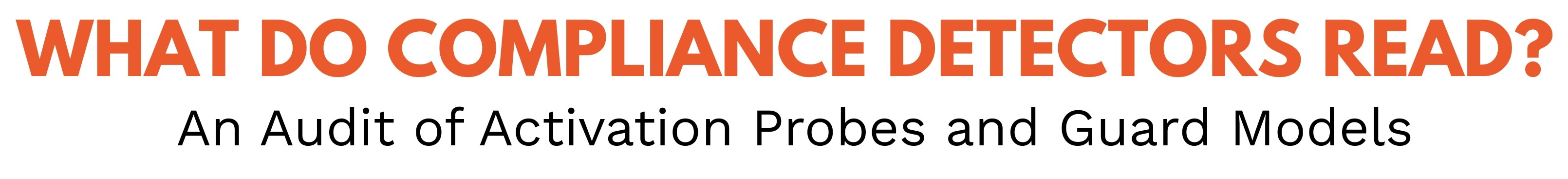}}
\author{
  Saisab Sadhu, Aadit Sengupta, Vinay Kumar Sankarapu, \\ Pratinav Seth \\
  \affiliation{Lexsi Labs} \\
  \texttt{saisab.sadhu@lexsi.ai}
}

\setabstract{%
Regulatory compliance monitoring in deployed language models is increasingly implemented as a legal and audit control, checking model outputs against written rules spanning data protection, healthcare, financial regulation, and platform policy. Such monitoring is meaningful only if a detector's verdict depends on the stated rule rather than on surface features of the scenario. We show this condition fails across the current class of compliance detectors, a failure we call rule blindness. Deleting, permuting, or substituting the governing rule leaves detection accuracy unchanged for every guard and activation probe we test, including a policy-conditioned guard that correctly cites the governing clause's position ninety one to ninety five percent of the time yet barely changes its verdict when that clause is swapped for its permissive counterpart. A purpose-built benchmark crossing two rules with two scenarios, so that neither alone predicts the label, confirms the failure under a design no prior benchmark rules out, and shows that step by step reasoning, not any fast detector we test, is what escapes it. Testing this at scale requires a detector cheap enough to run across many guards and benchmarks without retraining, so we introduce the Internal Compliance Score, a training-free activation readout calibrated from ten labelled pairs and scored by a single projection. We hold it to the same scrutiny as the guards it is compared against, and a pre-registered criterion for beating trivial baselines is not met, and a bag-of-words model matches its pooled generalisation exactly. It remains useful because it is inexpensive, letting us audit four deployed guard models, an 8B zero-shot judge, and thirteen benchmarks with no prior guard comparison, and it raises the mechanically verified pass rate when used to rank candidate responses, though an adaptive white-box attack removes this gain. We release the counterfactual protocol, including the crossed-rule benchmark, so rule blindness can be tested for in future probe and guard claims.
}

\setkeywords{compliance monitoring, activation probing, guard models, representation engineering, rule blindness}

\runningtitle{What Do Compliance Detectors Read? An Audit of Activation Probes and Guard Models}

\begin{document}
\maketitle

\section{Introduction}
\label{sec:introduction}
Deploying a language model in a regulated setting requires checking outputs against written rules, and this checking is increasingly performed by a guard model, a classifier fine-tuned on a large labelled corpus and run as an additional forward pass \citep{inan2023llamaguard,han2024wildguard,qwen2025qwen3guard}. Consider the storage-limitation principle in the General Data Protection Regulation (GDPR), which limits retention of personal data to what is necessary for the purpose of its collection \citep{gdpr2016}. A compliance monitor might operationalise this principle as a concrete rule, deletion within ninety days, and check it against a case in which data was retained for four hundred days. A deployed guard correctly flags the violation, yet flags the identical case even when that rule is replaced with an unrelated one or removed outright, so its verdict tracks the scenario rather than the governing rule. The same requirement recurs across data protection, healthcare privacy, financial regulation, and platform content policy alike, and the twenty regulatory domains evaluated in this paper span exactly this range, so the failure is not confined to one jurisdiction or industry.

We term this failure rule blindness, and show it is not confined to a single system, since deleting, permuting, or substituting the governing rule leaves detection accuracy unchanged for every guard model we test and for a training-free activation probe of our own. The guards divide into two kinds and both fail, for different reasons. Fixed-taxonomy safety classifiers such as Llama Guard~3 and Qwen3Guard never receive the rule, so they cannot track it by construction and their failure is a scope limitation rather than evidence about rule-conditioned systems. Policy-conditioned guards do receive it: LPG reads the rule through its own documented custom-policy channel, correctly cites the governing clause, and still barely changes its verdict when that clause is swapped for its permissive counterpart. The failure therefore spans the one-pass detector families we test, conditioned and unconditioned alike, rather than any particular architecture or training regime. An identical pattern has been reported independently for activation probes applied to safety classification \citep{siren2026,schwarz2026entanglement} (full positioning in the paper's Extended Related Work section (\S\ref{sec:discussion-related})), suggesting it is not specific to regulatory compliance. This matters because compliance monitoring is increasingly adopted as a legal and audit control across these sectors, and a detector that cannot distinguish a scenario from the rule that governs it offers only the appearance of rule-specific assurance, regardless of which rule is in force.

Establishing rule blindness at the scale of an entire field, across many guard models, activation probes, and benchmarks, requires a detector inexpensive enough to be run repeatedly without retraining. Guard models cannot meet this requirement, since fine-tuning demands a large labelled corpus, inference demands a complete additional forward pass, and the resulting classifier remains frozen and cannot track a monitored model that is later fine-tuned. We address this requirement by constructing the Internal Compliance Score (ICS), a difference-of-means readout in the lineage of representation engineering, the refusal direction, and mass-mean probing \citep{zou2023representation,arditi2024refusal,marks2023geometry,burns2023discovering}, applied here to written rules rather than safety labels. ICS reads the activations of the monitored model directly rather than querying a separate network, is calibrated from ten labelled pairs, and is scored by a single dot product, allowing the rule-blindness audit to be conducted across deployed guards (Llama Guard~3, WildGuard, Qwen3Guard, HarmBench \citep{mazeika2024harmbench}, SIREN \citep{siren2026}, GLiGuard, and Latent Policy Guard \citep{li2026lpg}), activation probes, an 8B zero-shot judge, and thirteen external benchmarks on which no prior guard comparison exists (\S\ref{sec:guardgap}); the nearest published neighbour is a compliance-framed whitening score with a different estimator and readout \citep{rachmil2025whitening}, positioned fully in the paper's Extended Related Work section (\S\ref{sec:discussion-related}).

ICS is evaluated against the same budget-matched selection null and pre-registered floor criterion applied to the guards it is compared against (\S\ref{sec:gate}). Among the methods compared here, ICS is the only one that is training-free, reads the monitored model directly, and recalibrates to a changed model at negligible cost, even though every method under test, ICS included, is rule-blind; the claim advanced for it is accordingly structural rather than one of superior accuracy. Whether such probes and their benchmarks can be trusted at all is a standing concern \citep{bailey2024obfuscated,mccoy2019hans}, positioned fully, including against dynamic policy-conditioned guards, in the paper's Extended Related Work section (\S\ref{sec:discussion-related}).

This paper makes four contributions, in that order of priority. 
\begin{itemize}
    \item \textbf{Method.} We introduce the Internal Compliance Score, a training-free activation readout that scores compliance risk directly from the monitored model, built from a difference of class means over a small labelled calibration set with no gradient step and one projection per case.
    \item \textbf{Evaluation.} We compare ICS against deployed guards, internal-representation probes, a gradient-based training-free detector, an 8B zero-shot judge, and lexical baselines, across compliance, safety, and held-out distributions, under budget-matched selection nulls, lexical floors, benign-input false-positive rates, and matched threshold budgets. The same pass audits the public compliance benchmarks themselves: four of seven are solvable by a policy-blind bag-of-words model, so they cannot adjudicate a rule-conditioned claim at all.
    \item \textbf{Characterisation.} Counterfactual rule interventions, and a purpose-built benchmark crossing rules with scenarios so that neither alone predicts the label, show that ICS and every efficient detector we test capture a broad violation signal more reliably than they compose the supplied rule with the scenario. We name this rule blindness, document it across every guard architecture and robustness check we run, and bound what any activation-based compliance readout can claim.
    \item \textbf{Application.} ICS ranks candidate responses well enough to raise the mechanically verified pass rate on IFEval by \rcpt{5.2} percentage points, while a white-box adaptive attack drives the same mechanism to its floor and so fixes a non-adversarial scope for it.
\end{itemize}

\section{Method: The Internal Compliance Score}
\label{sec:method}

\begin{figure*}[t]
\centering
\includegraphics[width=\textwidth]{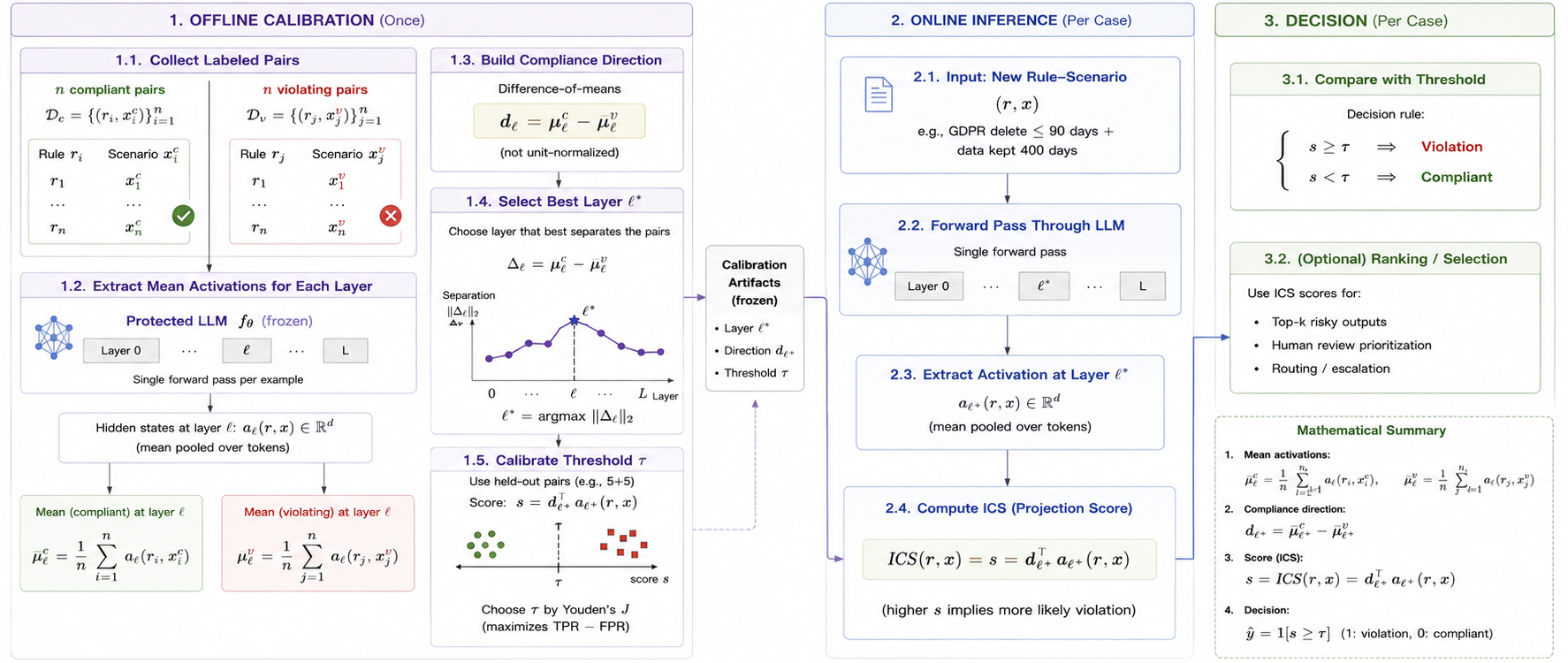}
\caption{\textbf{The full pipeline, build to score to decide.} Offline calibration (left) fits the direction and calibrates the anchor layer and threshold on two disjoint held-out slices, formalised opposite, before any test case is touched. Online inference (centre) costs one forward pass, one activation extraction, and one dot product. The decision stage (right) thresholds the score or, where useful, ranks or routes on it. Total cost is $2n$ forward passes to calibrate and one to score, no gradient step anywhere. Omitted for space: every scored arm is also checked against a budget-matched null, a lexical floor, and, where applicable, the pre-registered gate.}
\label{fig:architecture}
\end{figure*}

Figure~\ref{fig:architecture} gives the full pipeline this section formalises. We build $d$ once from ten pairs, calibrate $\ell^\ast$ and $\tau$ on two disjoint held-out slices, then score any new case at the cost of one dot product, with every reported number checked against a null, a floor, and the pre-registered gate. Let $M$ be a decoder transformer with $L$ blocks and residual width $d$. Let $a_\ell(x) \in \mathbb{R}^d$ denote the layer-$\ell$ residual-stream activation at the final token of text $x$, with the token position and site ablated in \S\ref{sec:ablations}. Given $n$ calibration pairs $\{(x_i^{+}, x_i^{-})\}_{i=1}^{n}$ of adherent and violating texts for a rule family $R$, the \emph{adherence direction} at layer $\ell$ is the normalized difference of class means,
\begin{equation}
\hat{d}_\ell \;=\;
\frac{\bar{a}_\ell^{+} - \bar{a}_\ell^{-}}
     {\lVert \bar{a}_\ell^{+} - \bar{a}_\ell^{-} \rVert + \varepsilon},
\qquad
\bar{a}_\ell^{\pm} = \tfrac{1}{n}\textstyle\sum_i a_\ell(x_i^{\pm}),
\label{eq:direction}
\end{equation}
and the score is the projection $\mathrm{ICS}_\ell(x) = \langle a_\ell(x), \hat{d}_\ell \rangle$.

The numerator of \eqref{eq:direction} points \emph{from} the violating prototype \emph{to} the adherent one, so \textbf{higher ICS means more adherent}, and every sign in the paper follows from that choice. Guardrail baselines use the opposite convention, an easy source of silent inversion we guard against explicitly with a $10^{-9}$ reference-implementation check.

Since $\langle a,\hat d\rangle = \lVert a\rVert\cos(a,\hat d)$, the raw score mixes angle with norm, and norm is a live confound (\texttt{length} alone scores \rcpt{0.566}), so we report the cosine variant beside it; the two tie, and $\lVert a\rVert$ alone is near chance (\S\ref{sec:ablations}). Equation~\eqref{eq:direction} with a midpoint threshold is nearest-centroid classification, Fisher's LDA under identity covariance, not parameter-free but requiring no optimizer and no hyperparameter sweep, though the full pipeline still spends labelled data on two validation slices, one to pick the layer and one to set the threshold, and at small $n$ approximately the regularized logistic solution (\S\ref{app:sampleeff}), which we treat as the informed ceiling. Each domain's pairs split $40/20/20/20$ into \textsc{train}, \textsc{val-layer}, \textsc{val-thresh}, and \textsc{test} (seed 42, grouped so no text straddles a boundary): the direction is fit on \textsc{train}, the anchor $\ell^{*}=\arg\max_\ell \mathrm{AUROC}_{\text{val-layer}}$ on the second (a single consensus layer costs $<0.01$), the threshold $\tau$ on the third, and \textsc{test} is touched once. We report AUROC as primary, since the equal-variance assumption behind $\tau$ fails, and $\mathrm{FPR}@95\%\mathrm{TPR}$ as the operating point; every headline finding survives re-expression under both and AUPRC (\S\ref{app:metric}), so none is an AUROC artifact.

The hypothesis is $I\!\left(a(x);Y\mid S(x)\right)>0$, that activations carry adherence signal beyond text surface $S(x)$, lexical content, sentiment, length. A benchmark that narrates the verdict makes $Y$ a function of $S(x)$ and any AUROC uninformative (\S\ref{sec:validity}), so every headline number is on outcome-ablated inputs and the headline statistic is the margin $\Delta=\mathrm{AUROC}_{\mathrm{ICS}}-\max_{b\in\mathcal B}\mathrm{AUROC}_b$ over trivial baselines, with paired-bootstrap CIs and BH correction. $\Delta\le 0$ is a reportable outcome, not a failure mode. Four statistical rules hold throughout. First, every metric appears with its exact null. Second, ceilings are one-sided Clopper--Pearson bounds, never $1.0$; raw perfect-separation point estimates are reported conservatively as $>0.999$. Third, best-of-$k$ is scored against its exchangeable null, $P(\arg\max\neq \text{first})=1-1/k$, the gain over a random-selection arm. Fourth, a permutation-null direction accompanies every detection table.

Unless noted, the model is Llama-3.2-1B-Instruct and the corpus is outcome-ablated OmniCompliance \citep{hu2026omnicompliance}, 20 domains, with the \texttt{Outcome:} field removed from every scored text (\S\ref{sec:validity}), same split above. The full model roster, baseline and floor pool, and statistics are in \S\ref{sec:setup}.

What ICS is calibrated on, and what it is tested on. Calibration and test are always disjoint, asserted throughout (\S\ref{sec:setup}), and only the calibration set's \emph{breadth} varies below. \emph{Per-distribution} calibration draws its pairs from the same distribution as the test, the strong in-domain case. \emph{Calibrate-once} withholds an entire distribution, building the direction from the others so the held-out one appears \emph{only} at test time, exactly as a deployed guard meets an input it was never trained on (Figure~\ref{fig:calibtest}).

\section{Evaluation Validity: What Compliance Benchmarks Can Measure}
\label{sec:validity}

This criterion determines which numbers in the paper are meaningful, so we state it before reporting any. OmniCompliance \citep{hu2026omnicompliance} structures each case as a rule, a scenario, and an \textsc{Outcome} field that states the regulatory result in natural language, for example ``fined EUR 725{,}000 for privacy violations,'' and the label is written into the input. On the as-published task the trivial baselines saturate, so no probe result on it is interpretable.

The obvious diagnosis, that a probe reads the narrated verdict, is \emph{refuted} by ablation. Removing the \textsc{Outcome} field and refitting barely moves detection (20-domain mean \rcpt{0.968} $\to$ \rcpt{0.952}): the field is sufficient for the label but not necessary, and the label is encoded redundantly across the scenario. The problem is therefore \textbf{degeneracy}, not contamination. When $Y$ is near-deterministic in the text surface $S(x)$, the conditional term $I(a(x); Y \mid S(x))$ has no room to be nonzero, bounding what \emph{any} method on the corpus can show, ours included. It is a property of provenance, not of one dataset: whenever a label is fixed before its prose is written, or recoverable from prose written afterwards, it survives in surface vocabulary. Consequently every detection number we report is on outcome-ablated inputs, carries its trivial-baseline rows and a permutation null, and bolds only the margin over the strongest \emph{named} baseline. We treat the ablated task as necessary but not sufficient, and the same audit applied to six further public compliance benchmarks leaves only two that can adjudicate a probe at all (Table~\ref{tab:benchmarks-main}); \S\ref{sec:degeneracy} adds benchmarks with mechanical rather than narrated ground truth. Splits are case-level, and we verified the leakage profile rather than assuming it (\rcpt{68\%} of test cases share a \texttt{source\_rule} with train, yet a rule-disjoint restriction leaves AUROC unchanged, \rcpt{0.962} vs.\ \rcpt{0.956}). This is a statement about what such corpora can \emph{evaluate}, not their construction quality.

\begin{table}[t]
\caption{\textbf{Four of seven public compliance benchmarks are lexically degenerate.} \emph{Lex.\ Floor}, strongest policy-blind bag-of-words model. \emph{Degenerate}, floor is high and the probe does not clear it, even for the human-expert-labelled one (AIReg-Bench). Full 14-benchmark table, Table~\ref{tab:benchmarks} in \S\ref{sec:degeneracy}.}
\label{tab:benchmarks-main}
\vspace{2mm}
\centering\small
\setlength{\tabcolsep}{9pt}
\fitwidth{%
\begin{tabular}{lrrrl}
\toprule
Benchmark & Lex.\ Floor & \textbf{ICS} & Budget Null & Verdict \\
\midrule
OmniCompliance & 0.896 & \textbf{0.952} & 0.714 & Degenerate \\
DynaBench & \textbf{0.982} & 0.784 & 0.576 & Degenerate \\
SafePyramid & 0.90+ & 0.871 & 0.673 & Degenerate \\
AIReg-Bench & \textbf{0.957} & 0.943 & 0.708 & Degenerate \\
CompliBench & 0.879 & 0.700 & 0.549 & Borderline \\
FlexBench & 0.802 & 0.832 & 0.603 & \textbf{Clean} \\
IFEval & \textbf{0.574} & 0.628 & 0.530 & \textbf{Clean} \\
\bottomrule
\end{tabular}
}
\end{table}

\section{Generalisation Beyond the Calibration Distribution}
\label{sec:guards}
\noindent A guard's value is that it is calibrated once and then handles whatever arrives, so the first question for a training-free readout is whether it survives leaving its calibration distribution. We answer that here and defer to \S\ref{sec:rule} the separate question of what the resulting score represents. We compare ICS against \emph{generative} guards (Llama Guard~3, Qwen3Guard-8B, WildGuard-7B, HarmBench), \emph{internal-representation} probes (SIREN, GLiGuard), an 8B zero-shot LLM judge, and Latent Policy Guard \citep{li2026lpg}, a learned latent reasoner sharing no method overlap with ICS but the same cheap-inference-time-adherence motivation, on the full outcome-ablated OmniCompliance benchmark, under every metric and on benign inputs.

We test whether a training-free direction is a \emph{calibrate-once guard competitive at near-zero cost}: generalisation, external sources, and efficiency support it; cross-family collapse, rule-blindness, and an adaptive attack bound it; in-domain accuracy is the control. Of the five detector types tested, \emph{zero} read the rule rather than a generic violation signal (\S\ref{sec:rule}) or hold across compliance and safety data (Table~\ref{tab:transfermat}). ICS's own cost advantage, zero trained parameters and roughly ten labelled pairs per source distribution ($\sim$100--160 pooled total, not ten total, \S\ref{sec:calib}) against every other arm's 1--8B parameters and $10^3$--$10^5$ pairs, is in Figure~\ref{fig:pareto}.

We pool compliant/violating and safe/unsafe pairs from \emph{eleven} distinct distributions (OmniCompliance, the three TRIDENT domains, and six safety benchmarks, Table~\ref{tab:lodo}), fit a \emph{single} difference-of-means direction, and test it \emph{leave-one-distribution-out}: calibrate with a distribution held out entirely, then score it. Mean AUROC on distributions never seen is \rcpt{0.728}, \rcpt{10}/11 above \rcpt{0.60}, clearing its budget-matched null (\rcpt{0.557}) at a \rcpt{$-0.12$} cost against per-distribution calibration (\rcpt{0.849}). The cross-family case is sharpest: with OmniCompliance held out entirely, trained only on TRIDENT and safety, the direction still reads regulatory compliance at \rcpt{0.783}, above the deployed guards on that distribution (Llama Guard~3 \rcpt{0.689}, SIREN \rcpt{0.697}). A pooled bag of words matches the pooled mean exactly (TF-IDF LODO \rcpt{0.728}), but that tie is an average: on the outcome-ablated compliance task itself, where lexical shortcuts are removed by design, the internal direction beats TF-IDF by \rcpt{$+0.26$}, and TF-IDF only draws level on distributions with residual surface cues, so ICS's case where it ties is structural, first-party, zero-parameter, and re-fittable rather than accuracy (\S\ref{app:matchedcalib}). Two scope conditions bound the result for a deployer. The \emph{direction}, not the operating threshold, is what transfers calibrate-once: a fixed cross-domain $\tau$ blows $\mathrm{FPR}@95$ up to \rcpt{77--94\%} (\S\ref{app:thresh}), so the threshold is re-calibrated per domain even as the direction is reused. And the generalisation is \emph{within-family}: withholding a whole family, all safety or all compliance, collapses cross-family transfer to chance (\rcpt{0.55}/\rcpt{0.48}, \S\ref{app:family}).

\begin{table}[t]
\caption{\textbf{Calibrate-once, leave-one-distribution-out} (AUROC$\uparrow$). \emph{Null}, budget-matched random directions; \emph{In-Dom.}, per-distribution ICS within this experiment's own protocol (an upper bound, not a second measurement of the \rcpt{0.952} headline figure). Clears its null (\rcpt{0.728} vs.\ \rcpt{0.557}) but ties TF-IDF in the mean.}
\label{tab:lodo}
\vspace{2mm}
\centering\small
\setlength{\tabcolsep}{9pt}
\fitwidth{%
\begin{tabular}{ll rrrr}
\toprule
Held-Out Distribution & Family & \textbf{ICS} & Null & TF-IDF & In-Dom. \\
\midrule
OmniCompliance & Compliance & 0.783 & 0.538 & 0.523 & 0.976 \\
TRIDENT-finance & Reg-QA & 0.737 & 0.587 & 0.836 & 0.837 \\
TRIDENT-law & Reg-QA & \textbf{0.549} & 0.555 & 0.865 & 0.777 \\
TRIDENT-med & Reg-QA & 0.917 & 0.627 & 0.937 & 0.916 \\
ToxicChat & Safety & 0.706 & 0.546 & 0.674 & 0.846 \\
OpenAI Mod. & Safety & 0.708 & 0.549 & 0.725 & 0.898 \\
Aegis~1.0 & Safety & 0.639 & 0.529 & 0.685 & 0.833 \\
Aegis~2.0 & Safety & 0.694 & 0.512 & 0.661 & 0.780 \\
WildGuard & Safety & 0.718 & 0.559 & 0.623 & 0.864 \\
PKU-SafeRLHF & Safety & 0.740 & 0.550 & 0.777 & 0.800 \\
BeaverTails & Safety & 0.820 & 0.574 & 0.703 & 0.810 \\
\midrule
\textit{Mean} & & \textbf{0.728} & 0.557 & 0.728 & 0.849 \\
\bottomrule
\end{tabular}
}
\end{table}

The complete calibrate-on-X, test-on-Y grid (Table~\ref{tab:transfermat}) confirms that block structure at every cell, and no single calibration ships everywhere: even the best single calibrator collapses on the opposite family. Calibrate on the family you will monitor.

A stricter check scores the three pool directions on \emph{seven} never-pooled external benchmarks sharing no row, benchmark, or collection process with any pool. The readout fires above chance on never-seen sources and the family structure mostly holds, but the compliance half holds only marginally: an external compliance set is read above chance by the \emph{shared} signal, not the compliance-specialised direction (Table~\ref{tab:external}).

\section{Comparison with Deployed Guards}
\label{sec:headtohead}

Calibrated on the test distribution directly (Table~\ref{tab:metrics}), ICS reaches AUROC \rcpt{0.952} on OmniCompliance and beats every deployed guard head-to-head (\rcpt{$+0.203$} over Llama Guard 3, \rcpt{$+0.255$} over SIREN, an MLP on the same activations, \rcpt{0.966}, and LPG-4B, \rcpt{0.943}, tie it). This is the in-domain case: ICS gets in-domain pairs while the guards run zero-shot out of their safety-training domain, so the leave-one-out number above, not this one, is the calibrate-once comparison the claim rests on.

A guard can screen a compliance case as an \emph{input} (Scenario in the prompt slot) or judge it as a \emph{response} (Scenario in the output slot against the Rule); the two are not interchangeable, so we run every generative guard in both modes and report both (\S\ref{app:guardio}). Response mode is stronger for all four guards on compliance, decisively for WildGuard, whose prompt-slot head is near-chance on \textsc{trident} and recovers only in response mode (\rcpt{0.785}$\to$\rcpt{0.875} AUROC on OmniCompliance). An earlier draft's per-metric tables here and in Table~\ref{tab:metrics} mixed a fixed prompt-slot mapping, the weaker of the two modes for Llama Guard~3, WildGuard, and Qwen3Guard, understating them by \rcpt{0.060}, \rcpt{0.090}, and \rcpt{0.023} AUROC respectively; every headline table now credits each guard its response mode uniformly, recomputed and verified per-domain-then-mean from the raw scores behind Table~\ref{tab:guardio}. The comparison the claim actually rests on, calibrate-once generalisation (Table~\ref{tab:lodo}), never used these rows either way. ICS sidesteps the mode choice, since it reads the activation over the case text itself.

At their shipped threshold, guards rarely flag a regulatory violation (recall \rcpt{0.08--0.22}), so native F1 is low against ICS's \rcpt{0.904} (Table~\ref{tab:metrics-main}); this is mostly a threshold artifact, not a capability gap. Given every guard the \emph{same} target-domain threshold budget ICS gets (native, ten pairs, the full budget, or a Platt/isotonic map, fit held-in and scored on a disjoint fold), the collapse recovers to F1 \rcpt{0.54}--\rcpt{0.80} (Table~\ref{tab:matchedcalib}). ICS's advantage is \emph{ranking}, not threshold access: its AUROC (\rcpt{0.952}) clears every guard's (\rcpt{0.735}--\rcpt{0.901}) budget-free, and at matched calibration it still leads every threshold metric (F1, FPR@95, TPR@10), so ranking, or the calibrate-once generalisation (Table~\ref{tab:lodo}), is the comparison that holds.

Two supporting results are summarised here and given in full in \S\ref{app:matchedcalib}. The training-free story is not idiosyncratic to difference-of-means: GradSafe, a gradient-based training-free detector, also beats every deployed guard on compliance once re-referenced with sixteen compliance pairs, and collapses to chance under its native safety reference, the same failure the safety-trained guards show. And because ICS reads the monitored model's own activations, it can be re-fit on that model after fine-tuning, where a shipped guard is a separate frozen network; a fresh ten-pair refit holds across nine models and multiple fine-tunes, AUROC within \rcpt{0.007} of base (\S\ref{app:matchedcalib}).

Guards are trained for safety, so we also run every arm on their home domain, the standard seven safety benchmarks, by Macro F1 (Table~\ref{tab:guards}), and on benign inputs (Table~\ref{tab:benign}). There they are strong on detection but pay in \emph{over-refusal}, WildGuard flagging \rcpt{75\%} of adversarially-benign OR-Bench prompts, which is what a comparison run only on regulatory compliance, the guards' non-native domain, would otherwise obscure. \S\ref{sec:rule} shows the whole class shares a rule-blind spot.

\begin{table}[t]
\caption{\textbf{Detector comparison, in-domain.} Native-threshold F1 (\emph{Native}) vs.\ ten-pair-budget F1 (\emph{Recal.}); guard AUROCs now uniformly credit each guard its stronger, response mode (\S\ref{sec:headtohead}, full metrics in Table~\ref{tab:metrics} and Table~\ref{tab:matchedcalib}).}
\label{tab:metrics-main}
\vspace{2mm}
\centering\small
\setlength{\tabcolsep}{9pt}
\fitwidth{%
\begin{tabular}{lrrr}
\toprule
Arm & AUROC & F1 (Native) & F1 (Recal.) \\
\midrule
\textbf{ICS} (ours) & \textbf{0.952} & \textbf{0.897} & \textbf{0.895} \\
Qwen3Guard-Gen (8B) & 0.908 & 0.158 & 0.804 \\
LLM judge (zero-shot, 8B) & 0.901 & 0.831 & 0.817 \\
WildGuard (7B) & 0.875 & 0.124 & 0.715 \\
HarmBench cls (7B) & 0.735 & 0.094 & 0.656 \\
Llama Guard 3 (1B) & 0.749 & 0.265 & 0.580 \\
\bottomrule
\end{tabular}
}
\end{table}



\section{What the Score Measures}
\label{sec:rule}

The sections above establish that ICS ranks compliance risk competitively and at low cost. None of it establishes \emph{what} the score is reading, and that is the question this section answers, for ICS \emph{and} for the guards it was compared against. The task, as stated, asks whether a given scenario violates \emph{that} rule, not some other one, so a detector at AUROC \rcpt{0.952} could be composing the rule with the scenario or could be recognising that the scenario sounds like a violation and ignoring the rule entirely. Conventional accuracy cannot separate the two. We test whether the probe uses the rule at all by refitting ICS from scratch under four corruptions of the rule block and one of the scenario (Table~\ref{tab:rule}). It does not, and the finding is not special to ICS: the same corruptions leave a 4B policy guard with an explicit policy channel unmoved (below), so a first-party probe and a shipped guard that is actually given the rule both read a generic violation signal rather than the rule-scenario relation. Nor is it an artifact of a benchmark whose label ignores the rule, since it survives the sharpest available \emph{counterfactual}, SafePyramid's surface-matched flip pairs, where the conversation is byte-identical within a pair and the policy alone flips the label (below).

\begin{table}[t]
\caption{\textbf{The rule is optional.} \emph{Sig.\ Drops}, BH-significant AUROC decreases vs.\ correct-rule (exact arithmetic, 20 domains). A wrong-domain rule carries zero class information yet costs nothing, and rule-only is the positive control. Replicates on four further models (Table~\ref{tab:rule-crossmodel}).}
\label{tab:rule}
\vspace{2mm}
\centering\small
\setlength{\tabcolsep}{9pt}
\fitwidth{%
\begin{tabular}{lrrrc}
\toprule
Condition & ICS (1B) & Null & $\Delta$ (Cond$-$Corr) & Sig.\ Drops \\
\midrule
Correct rule & 0.9519 & 0.714 & --- & --- \\
\textbf{No rule at all} & \textbf{0.9549} & 0.719 & $\mathbf{+0.003}$ & \textbf{0/20} \\
Shuffled rule & 0.9548 & 0.713 & $+0.003$ & 0/20 \\
Wrong-domain rule & 0.9554 & 0.712 & $+0.003$ & 0/20 \\
\midrule
Rule only, no scenario & 0.5627 & 0.527 & $-0.389$ & \textbf{19/20} \\
\midrule
\multicolumn{5}{l}{\textit{TF-IDF, same conditions (null $=0.5$, identical by construction)}} \\
Correct rule & 0.8956 & 0.500 & --- & --- \\
\textbf{No rule at all} & \textbf{0.9308} & 0.500 & $\mathbf{+0.035}$ & 0/20 \emph{(13 gains)} \\
Rule only & 0.5881 & 0.500 & $-0.308$ & 19/20 \\
\bottomrule
\end{tabular}
}
\end{table}

The rule is unnecessary, and it is not an artifact. Deleting the rule moves mean AUROC by \rcpt{$+0.003$}, a slight \emph{help}, with \emph{none} of 20 domains showing a BH-significant drop and \emph{none} a gain, symmetric noise rather than an underpowered near-miss. Shuffling the rule and substituting a different domain's rule give the same answer, and it is not length or truncation: the three conditions are length-identical, no test text nears the 512-token cap, and the pipeline reproduces the published gate bit-exactly before any condition runs. Nor is it an artifact of one model, one benchmark, one rule source, or our particular readout. Table~\ref{tab:robust} collects every check we ran against those possibilities; all of them leave the finding intact, and the two positive controls confirm the test has power to detect change.

A fixed-scenario counterfactual isolates the readout from the model, but the natural design has a confound a reviewer specifically flagged: holding one scenario fixed and flipping only the rule means rule polarity alone, permissive versus prohibitive, determines the label throughout the set, so a detector that has merely learned to read polarity, not the rule-scenario interaction, could pass without doing anything relational. We resolve this with a crossed design that removes the confound by construction rather than by argument. Each of \rcpt{200} templates across \rcpt{8} regulatory domains supplies two rules and two scenarios crossing into four labelled cells, rule R0 with scenario Sa a violation, R0 with Sb compliant, R1 reversing both, so neither the rule text nor the scenario text alone predicts the label, each appearing exactly once with each outcome across the \rcpt{800} rows. Labels come from a numeric threshold comparison, computed mechanically, never judged, and the no-shortcut guarantee is verified rather than assumed: a TF-IDF classifier trained on the rule alone, the scenario alone, or the two concatenated scores exactly chance, \rcpt{0.500} AUROC, group-disjoint five-fold cross-validation by quadruple (\S\ref{app:crossed}).

On this genuinely unconfounded set, every cheap detector we test sits at or barely above chance, and the two detector classes built specifically to escape rule blindness do not (Table~\ref{tab:crossed}). LPG was designed to condition on the active policy rather than surface priors, and its own violated-policy-removal probe already showed two baseline dynamic guards flipping to safe only 64\% and 36\% of the time when the violated policy is removed rather than reliably, evidence the authors read as guards conditioning on positional and content priors rather than the supplied clauses \citep{li2026lpg}, their figures, not ours. LPG does not escape on our crossed set either. Refitting a probe directly on the crossed data, the strongest possible case for a shallow readout, recovers a real but weak scale-dependent signal, Qwen3-8B topping out at \rcpt{0.670} AUROC and \rcpt{11.0}\% quadruple exact match against a \rcpt{6.25}\% chance rate, not a solved task. One configuration does substantially better: the identical local judge, given the same rule and scenario but allowed to reason step by step rather than forced to a single token, reaches \rcpt{0.849} AUROC and \rcpt{74.4}\% quadruple exact match on a seeded \rcpt{40}-quadruple subsample (Wilson CI \rcpt{58.9}--\rcpt{85.4\%}). That control is neither compute-matched against the forced-choice row nor run over the full set, so it establishes that the benchmark is solvable, not that step-by-step reasoning is necessary or sufficient for it. What it does rule out is a degenerate design, leaving the failure specific to detectors that read a generic violation signal in one forward pass rather than explicitly composing the rule and the scenario.

\begin{table}[t]
\caption{\textbf{Confound-free crossed benchmark: every cheap detector is at chance; chain-of-thought is the one configuration that moves.} Chance is 6.25\% quadruple exact match, 0.500 AUROC. Chain-of-thought is a positive control (40/200 quadruples, one prompt, one seed, not compute-matched against forced-choice; Wilson CI \rcpt{58.9}--\rcpt{85.4\%}), so it shows the benchmark is solvable, not that reasoning is necessary or sufficient. Full detail in \S\ref{app:crossed}.}
\label{tab:crossed}
\vspace{2mm}
\centering\small
\setlength{\tabcolsep}{9pt}
\fitwidth{%
\begin{tabular}{@{}lrr@{}}
\toprule
Detector & AUROC & Quad.\ Exact (\%) \\
\midrule
TF-IDF (rule, scenario, or both) & 0.500 & 0.0 \\
Frozen ICS (1B/8B/Qwen3-8B) & 0.50--0.54 & 0.0--3.5 \\
Best refit (Qwen3-8B MLP) & 0.670 & 11.0 \\
Llama Guard 3 / Qwen3Guard & 0.50--0.52 & 0.0--0.5 \\
LPG-4B (rule as custom policy) & 0.535 & 1.0 \\
Rachmil whitening (their guards) & 0.498 & 0.0 \\
Forced-choice judge (1 token) & 0.507 & 0.0 \\
\textbf{Same judge, chain-of-thought} & \textbf{0.849} & \textbf{74.4} \\
\bottomrule
\end{tabular}
}
\end{table}



A companion, earlier fixed-scenario counterfactual replicates the same qualitative pattern with a caveat the crossed design above resolves (full detail, cross-model replication, and an independent frontier-judge cross-check against Haiku 4.5, GPT-4o, and Gemini 2.5 Flash under the local judge's exact prompts, in \S\ref{app:fixedscenariocf} and \S\ref{app:frontierjudge}, including the frontier-judge validation figure there).
\begin{table}[t]
\caption{\textbf{Rule blindness across every robustness check.} Each row refits ICS from scratch under the stated condition against correct-rule; \emph{Holds} means the no-rule result is unchanged. The two positive controls show the test has power: removing the \emph{scenario} collapses detection, and the topical rule axis is decodable but label-orthogonal. Full derivations and replication detail throughout this appendix.}
\label{tab:robust}
\vspace{2mm}
\centering\small
\setlength{\tabcolsep}{9pt}
\fitwidth{%
\begin{tabular}{@{}llr@{}}
\toprule
Robustness check & Result & Holds? \\
\midrule
\multicolumn{3}{@{}l}{\textit{Is the no-rule result an artifact of one model?}}\\
Llama-3.2-3B, Llama-3.1-8B & 0/20 sig.\ drops & yes \\
Qwen2.5-7B, Qwen3-8B & 2/20, negligible effect & yes \\
LoRA verdict-SFT ($0.67{\to}0.93$ acc.) & 0/6 sig.\ drops & yes \\
\midrule
\multicolumn{3}{@{}l}{\textit{Of the benchmark, or of machine-written rules?}}\\
\textsc{trident} (real CFA/ABA/AMA codes) & $\Delta<0.001$, 0/3 & yes \\
SafePyramid surface-matched pairs & 0.5000 vs.\ null 0.519 & yes \\
\midrule
\multicolumn{3}{@{}l}{\textit{Of reading the rule at the wrong place?}}\\
LPG, permissive counterpart swapped in & flips 7.1\% of cases & yes \\
Refusal-orthogonalised contrastive fit & 0.605, below null 19/20 & yes \\
\midrule
\multicolumn{3}{@{}l}{\textit{Positive controls (these must move, and do)}}\\
Rule only, no scenario & 0.56--0.57 & --- \\
On- vs.\ off-topic rule (topical axis) & decodable 0.90, label-orthogonal & --- \\
\bottomrule
\end{tabular}
}
\end{table}

The most direct repair also fails: building the direction so that it \emph{must} attend to the rule, calibrating from badness-matched rule-varying contrasts and orthogonalising against the refusal axis after conditional activation steering \citep{lee2024cast}, lands at \rcpt{0.605}, below its null in \rcpt{19/20} domains, and deleting the rule still costs nothing. A shipped policy-conditioned guard fares no better: swapping the violated rule for its exact permissive counterpart, scenario and decoy policies fixed, flips LPG's verdict on only \rcpt{7.1\%} of cases even though it cites the governing clause's list position \rcpt{91}--\rcpt{95\%} of the time (\S\ref{app:lpgremoval}). Citing the rule and adjudicating against it are separable, and these detectors do the first.
ICS is therefore better interpreted as a broad compliance-risk direction than as a rule-specific adjudication direction. Our geometry section could not decide between these two readings within a single-template benchmark. This measurement does.


\section{Score-Guided Candidate Selection}
\label{sec:results}\label{sec:selection}

The strongest practical result is on substantively different generated responses, not menu labels, and is mechanically verified: an ICS-guided pick of instruction-following responses beats random by \rcpt{+5.2}pp against a mechanical verifier (\rcpt{541} cases), scaling with within-prompt detection AUROC. A frozen direction, the same one, no re-fitting, also beats random by \rcpt{+11.5}pp on LLM-judged regulatory advisory responses across six domains, surviving a coherence control that rules out quality-selection on the judged arm (\S\ref{app:contentsel}). Across a mechanical verifier and a judge, on format and on regulation, the gain is real and modest. It holds in non-adaptive settings only: a white-box suffix optimised against the direction drives the same selection to its floor (\S\ref{app:attack}), so this is not robust enough for standalone adversarial enforcement; the supported use is score-guided ranking under trusted or non-adaptive candidate generation, not adversarial enforcement.

As a controlled sanity check, the same mechanism is tested on a classification corpus, OmniCompliance, where best-of-$k$ selection ($k{=}5$, pure temperature sampling, ablated inputs) returns the candidate whose ICS-predicted verdict is correct (Table~\ref{tab:selection}), capturing \rcpt{64--79\%} of the oracle headroom in every one of six models from 1B to 9B, significant and positive with zero significant negative domains anywhere. This is \emph{verdict}-selection over sampled menu labels, not content-selection, so the headline headroom number is the more modest generated-response result above, not this one. The absolute benefit is governed by the generator's blind accuracy: gain over the random arm tracks the \emph{available} headroom (oracle $-$ random) with slope \rcpt{0.79}, intercept \rcpt{$-0.003$}, and $R^2 = \rcpt{0.94}$ across \rcpt{120} model-domain points (mean \rcpt{$+0.27$} where blind accuracy $<0.70$ versus \rcpt{$+0.02$} where $>0.85$; Figure~\ref{fig:headroom}). ICS-guided selection also beats the same-model zero-shot verdict logit, a readout needing no direction, at \rcpt{0.872 vs.\ 0.703} on the 1B, McNemar $p = 1.2\times10^{-95}$, \emph{all} 20 domains, with candidate diversity verified directly (both verdicts present in \rcpt{88.9\%} of cases), confirming the direction itself does the work rather than the sampling.

\begin{table}[t]
\caption{\textbf{Verdict-selection sanity check, best-of-5} (outcome-ablated OmniCompliance verdict accuracy). \emph{Blind}, no selection; \emph{Random}/\emph{ICS-Guided}/\emph{Oracle}, best-of-5 by random, ICS, and true-verdict pick; headroom $=$ (guided$-$random)/(oracle$-$random). Absolute gain shrinks on stronger generators as their room shrinks; the recovered fraction (64--79\%) does not.}
\label{tab:selection}
\vspace{2mm}
\centering\small
\setlength{\tabcolsep}{9pt}
\fitwidth{%
\begin{tabular}{lrrrrr}
\toprule
Generator & Blind & Random & \textbf{ICS-Guided} & Oracle & Headroom \\
\midrule
Llama-3.2-1B & 0.569 & 0.541 & \textbf{0.872} & 0.961 & 78.8\% \\
Llama-3.2-3B & 0.771 & 0.773 & \textbf{0.889} & 0.946 & 66.7\% \\
Mistral-7B & 0.775 & 0.770 & \textbf{0.795} & 0.805 & 71.8\% \\
Llama-3.1-8B & 0.783 & 0.781 & \textbf{0.889} & 0.945 & 66.1\% \\
Gemma-2-9B & 0.797 & 0.793 & \textbf{0.832} & 0.848 & 70.4\% \\
Qwen2.5-7B & 0.843 & 0.842 & \textbf{0.862} & 0.873 & 63.9\% \\
\bottomrule
\end{tabular}
}
\end{table}

\section{Scope and Limitations}
\label{sec:limitations}
Five limits bound the headline claims. ICS requires first-party activation
access, an advantage for self-monitoring, not third-party auditing. Its
direction transfers within related task families, collapsing across
compliance and safety data (\rcpt{0.55}/\rcpt{0.48}, \S\ref{app:family}),
and its threshold needs local recalibration regardless. The score captures a
broad compliance-risk signal rather than reliable rule-specific composition.
A trained MLP or logistic probe on the same activations is slightly more
accurate (\rcpt{0.966} vs.\ \rcpt{0.952}) at every calibration size tested, a
cost trade rather than an accuracy one (\S\ref{app:sampleeff}). Finally, the
readout is serialisation-sensitive, collapsing to \rcpt{0.668} when the
scored text ends on the rule rather than the case, and an adaptive
white-box attack defeats it as a standalone enforcement control
\citep{bailey2024obfuscated}. Further scope conditions in
\S\ref{app:pitfalls}.

\section{Conclusion}
\label{sec:conclusion}
The Internal Compliance Score is a training-free readout of the monitored
model's own activations, a difference of means over a small calibration set
scored by one projection per case, no gradient step. It ranks compliance
risk competitively in domain, keeps useful signal on distributions withheld
from calibration (leave-one-out AUROC \rcpt{0.728} against a \rcpt{0.557}
null, worst case \rcpt{0.549}), and re-fits after fine-tuning, which a
separate guard cannot. Counterfactual testing bounds what the score
represents. It reads as a compliance-risk direction rather than a
rule-specific adjudication direction, since deleting or substituting the
rule leaves it, and every guard and probe tested, unchanged. A bag of words
matches its pooled generalisation, transfer between compliance and safety
data is weak, and an adaptive attack removes its selection gain. ICS is a low-cost method for monitoring, triage, and candidate ranking, released with the counterfactual protocol that establishes those limits.
\clearpage
\appendix

\FloatBarrier
\section{Experimental Setup}
\label{sec:setup}

\paragraph{Models.} Table~\ref{tab:setup} lists the roster. All extraction is \texttt{bfloat16}, last-token, residual stream. Layers are indexed $0,\ldots,L$, so the anchor search over the layer-validation slice ranges over $L{+}1$ positions including the embedding (17 for the 1B), and the pipeline reproduces the stored gate receipts bit-exactly before any new arm runs. Detection and geometry use Llama-3.2-1B-Instruct, and the remaining arms replicate detection and, per the paper's Score-Guided Candidate Selection section, selection.

\paragraph{Compute budget.} All detection, gate, and cross-model arms (1B--72B, twelve models) run on a single GPU per arm, the two frontier-scale (70B/72B) rows in int8 (\S\ref{sec:models}); no arm requires multi-GPU sharding. Calibration is ten to two hundred labelled pairs per domain and one forward pass per case, so the full 20-domain, twelve-model detection sweep is on the order of low-thousands of forward passes, not a training run. The adaptive white-box attack (\S\ref{app:attack}) is the single most compute-intensive component, a 40-step GCG \citep{zou2023adversarial} search per prompt over 20 stratified prompts; every other reported number is inference-only. Total compute is released with the evaluation protocol.

\begin{table*}[htbp]
\caption{\textbf{Twelve models, four architecture families, 1B to 72B.} All extraction \texttt{bfloat16} except the 72B arm (int8; see the 72B Arm methodology note in Cross-Model Replication, \S\ref{sec:models}). $^\dagger$Run in an isolated environment (\texttt{transformers}~5.14). A plain \textbf{---} means not applicable to that row; a separate \textbf{---} in the $d$ column means not required for this paper's experiments.}
\label{tab:setup}
\vspace{2mm}
\centering\small
\setlength{\tabcolsep}{9pt}
\fitwidth{%
\begin{tabular}{llrrl}
\toprule
Model & Family & $L{+}1$ & $d$ & Role \\
\midrule
Llama-3.2-1B-Instruct & Llama & 17 & 2048 & primary (detection, geometry, keystone) \\
Llama-3.2-3B-Instruct & Llama & 29 & 3072 & detection, rule-blindness \\
Llama-3.1-8B-Instruct & Llama & 33 & 4096 & detection, rule-blindness, keystone, geometry \\
Gemma-2-9B-it & Gemma & 43 & 3584 & detection \\
Mistral-7B-Instruct-v0.3 & Mistral & 33 & 4096 & detection \\
Qwen2.5-7B-Instruct & Qwen & 29 & 3584 & detection, rule-blindness \\
Qwen3-8B & Qwen & 37 & 4096 & detection, rule-blindness, keystone, geometry \\
Llama-3.3-70B-Instruct & Llama & 81 & 8192 & detection (cross-model replication, 70B arm) \\
Qwen3.5-4B / 9B$^\dagger$ & Qwen & 33 & 2560 / 4096 & detection \\
Qwen3.5-27B$^\dagger$ & Qwen & 65 & 5120 & detection \\
\textbf{Qwen2.5-72B-Instruct (int8)} & Qwen & \textbf{81} & \textbf{8192} & \textbf{detection (frontier-scale arm)} \\
\bottomrule
\end{tabular}
}
\end{table*}

\paragraph{Data.} The primary corpus is outcome-ablated OmniCompliance \citep{hu2026omnicompliance}, 20 domains, with the \texttt{Outcome:} field removed from every scored text (the paper's Evaluation Validity section). Detection fits the direction on the 40\% train slice of each domain's \rcpt{73--500} pairs and reports AUROC on its \rcpt{16--100}-pair test slice. Selection uses its own generation sets of \rcpt{200} cases per domain (the paper's Score-Guided Candidate Selection section). We additionally run the gauntlet on six external compliance benchmarks and seven safety benchmarks (Table~\ref{tab:benchmarks}). Splits are 40/20/20/20 (train, layer-val, threshold-val, test), seed 42, grouped so no unit sharing text straddles a boundary.

\paragraph{Baselines and the floor pool.} The gate (\S\ref{sec:gate}) is defined against a \emph{pre-registered} floor set, TF-IDF, keyword, length, VADER, the same-model Yes/No verdict logit, the SST-2 sentiment direction, and an Arditi-style refusal direction fit with zero compliance supervision \citep{arditi2024refusal}, fixed before any experiment ran. The \rcpt{11/20} verdict is the margin over the per-domain maximum of this exact set. Tier-2 estimators (logistic, CCS, shrinkage-LDA, whitened difference-of-means, a one-hidden-layer MLP) and Tier-3 deployed systems (Llama Guard 3, Qwen3Guard-8B, SIREN, GLiGuard, Latent Policy Guard, an 8B LLM judge) are compared in the paper's Generalisation Beyond the Calibration Distribution section. Every selected direction is additionally scored against a \emph{budget-matched selection null}, 200 random unit directions pushed through exactly that arm's hyperparameter search.

\paragraph{Statistics.} We use paired bootstrap with the pair as the resampling unit ($B{=}2000$), Benjamini--Hochberg across the 20 domains with thresholds in exact rational arithmetic (\S\ref{sec:gate} reports a verdict that turns on this), and Clopper--Pearson bounds where a cell saturates, with anchor-layer stability read from $5{\times}2$ seed-replicate grids. Every AUROC passes through a wrapper asserting agreement with a reference implementation to $10^{-9}$, a check that caught a sign inversion during development (the paper's Method section). Provenance-stamped JSONs (model, seed, commit, dtype, pooling, \texttt{max\_length}) are released with the code.

\FloatBarrier
\section{The Pre-Registered Gate}
\label{sec:gate}

\noindent\textbf{A pre-registered floor test, reported in full.} Detection is cheap, but is its \emph{absolute} margin over trivial baselines real? We committed the rule before any experiment ran. Ablated ICS must beat the maximum trivial floor with a paired-bootstrap 95\% CI excluding zero, BH-corrected, in at least 12 of 20 domains. With the \emph{full floor pool} (this section's complete, corrected set of trivial baselines), including the corrected SST-2 sentiment direction, the maximum floor in 14/20 domains, ICS clears it in \rcpt{\textbf{11/20}}, just short of the pre-registered bar, an honest limit on the \emph{absolute} detector we report rather than bury (the deployable claims, namely selection and the guard comparison, do not rest on it). Mean ICS is \rcpt{0.952} against a mean per-domain maximum floor of \rcpt{0.925} (Table~\ref{tab:gate}). The strongest single floor arm is the SST-2 sentiment direction (mean \rcpt{0.914}, the maximum in 14/20 domains), with TF-IDF at \rcpt{0.896}.
\begin{table*}[htbp]
\caption{\textbf{The pre-registered gate, outcome-ablated} (8 of 20 domains, full table in \S\ref{app:gate}). Floor is the per-domain \emph{maximum} over trivial baselines, and an SST-2 sentiment direction fit on movie reviews is that maximum in \rcpt{14/20}. The rule, committed before any experiment, needs $\Delta>0$ with a BH-corrected paired-bootstrap CI excluding zero in $\ge$12/20. It reaches \textbf{11/20}.}
\label{tab:gate}
\vspace{2mm}
\centering\small
\setlength{\tabcolsep}{9pt}
\fitwidth{%
\begin{tabular}{lrrrlrc}
\toprule
Domain & $n$ & \textbf{ICS} & Max.\ trivial floor & (Which floor) & $\Delta$ & Gate \\
\midrule
Data Act & 100 & 0.964 & 0.894 & \textsc{sst2} & $+0.070$ & $\checkmark$ \\
Policy: Google & 100 & 0.970 & 0.919 & \textsc{sst2} & $+0.051$ & $\checkmark$ \\
HIPAA & 100 & 0.983 & 0.941 & \textsc{sst2} & $+0.042$ & $\checkmark$ \\
EU AI Act & 100 & 0.986 & 0.938 & \textsc{sst2} & $+0.048$ & $\checkmark$ \\
\midrule
Fin.\ crypto & 100 & 0.997 & 0.985 & \textsc{sst2} & $+0.012$ & -- \\
GDPR & 100 & 0.863 & 0.862 & \textsc{sst2} & $+0.001$ & -- \\
Edu: online & 72 & 0.997 & 0.999 & \textsc{sst2} & $-0.003$ & -- \\
Edu: discrim. & 16 & 0.875 & 0.887 & \textsc{tfidf} & $-0.012$ & -- \\
\midrule
\textit{Mean, 20 domains} & & \textit{0.952} & \textit{0.925} & & \textit{+0.027} & \textbf{11/20} \\
\bottomrule
\end{tabular}
}
\end{table*}

The verdict is 11/20 rather than 13/20 for two reasons that are themselves findings about evaluation practice. An earlier computation silently dropped the sentiment floor through a name-prefix filter, and one domain's pass rested on a floating-point BH threshold ($0.03{\times}20/12 = 0.0499\ldots$) that fails under exact arithmetic. We report all thresholds exactly, and at meta-analytic boundaries, representation error flips verdicts. Table~\ref{tab:crossmodel}'s cross-model gate column deliberately reuses a \emph{cheap floor} (the pre-correction, partial baseline set), not this section's full floor pool, to keep all twelve arms on an identical comparison. Its \textbf{13} for this same model is that cheap-floor figure, not a second disagreeing measurement of this section's full-floor \rcpt{11/20}, and the table's caption reconciles the two.

The failure pattern is structured but not clean, and we state both halves. All eleven passes are full-size ($n{=}100$) mid-difficulty domains. Of the nine failures, three are small-$n$ ($n \le 36$), two are ceiling-saturated (floor $\ge 0.98$), and two lose their pass to the corrected sentiment floor closing within 0.002--0.012, but two (GDPR, foundational rights) fail at $n{=}100$ against mid floors, so low power and high floors do not explain every miss. Where the margin is measurable at all, it is small (mean $+0.038$ over the sentiment direction).

\subsection{Full-Detail Tables Referenced from the Main Text}
\label{app:guardtables}
Five tables from the paper's What the Score Measures section, the paper's Generalisation Beyond the Calibration Distribution section, and the paper's Comparison with Deployed Guards section in full. Each is summarised in the main text and referenced from there; this section exists so the underlying numbers are checkable in full, not because a reader needs them to follow the argument.

\begin{table}[t]
\caption{\textbf{The no-rule condition (the paper's rule-is-optional table) replicated on four further models, across two families.} Llama holds fully (0/20 significant drops on both arms), Qwen shows the same small hedge on both models, an architecture-consistent pattern, not a single-checkpoint artifact.}
\label{tab:rule-crossmodel}
\vspace{2mm}
\centering\small
\setlength{\tabcolsep}{9pt}
\fitwidth{%
\begin{tabular}{lrrrc}
\toprule
Model & Correct & No Rule & $\Delta$ & Sig.\ Drops (Gains) \\
\midrule
Llama-3.2-1B (main text) & 0.9519 & 0.9549 & $+0.003$ & 0/20 \\
Llama-3.2-3B & 0.9580 & 0.9565 & $-0.0015$ & 0/20 \\
Llama-3.1-8B & 0.9624 & 0.9604 & $-0.0020$ & 0/20 (1) \\
Qwen2.5-7B & 0.9631 & 0.9604 & $-0.0027$ & \textbf{2/20} (0) \\
Qwen3-8B & 0.9528 & 0.9493 & $-0.0035$ & \textbf{2/20} (1) \\
\bottomrule
\end{tabular}
}
\end{table}

\subsection{The Fixed-Scenario Counterfactual, in Full}
\label{app:fixedscenariocf}
The sharpest fixed-scenario test fixes the scenario and flips the label with the rule, which no public benchmark does. We construct \rcpt{208} such pairs across \rcpt{8} regulatory domains, one real scenario, byte-identical across the two members, paired with a permissive rule (compliant) and a prohibitive one (violation), condition-based and length-matched so that no lexical polarity survives (scenario-only and cheap rule-cue baselines sit at \rcpt{0.49} and \rcpt{0.71}). On this set the deployed ICS direction reads the flip at chance (\rcpt{0.502}, within-pair \rcpt{0.587}), though it separates OmniCompliance at \rcpt{0.964} from the same layer. A direction \emph{fit on the flips} recovers them at \rcpt{0.736}, and a logistic or small MLP probe on the same last-token activation reaches \rcpt{0.874} and \rcpt{0.887}. As the paper's What the Score Measures section notes, this design cannot distinguish genuine rule-scenario reading from a probe that has merely learned rule polarity, since rule polarity alone predicts the label throughout this set, so these fit-on-the-flips numbers are an upper bound on relational reading, not proof of it, resolved by the crossed benchmark of \S\ref{app:crossed}. The chance-level readout of the \emph{deployed} (not fit-on-the-flips) direction replicates across model scale and family regardless of that caveat, since the deployed direction never sees the flips. Llama-3.1-8B-Instruct reads the identical flip set at \rcpt{0.504} and Qwen3-8B at \rcpt{0.489}, both indistinguishable from the 1B arm's \rcpt{0.502}, while the same fit-on-the-flips recovery holds on both. Logistic and MLP probes on Llama-3.1-8B reach \rcpt{0.943} and \rcpt{0.943}, and on Qwen3-8B \rcpt{0.905} and \rcpt{0.911}. The deployed-detector failure is class-wide. On this set the deployed probe (\rcpt{0.502}) and both fixed-taxonomy guards, which never receive the rule, sit at chance (Llama Guard 3 \rcpt{0.497}, Qwen3Guard \rcpt{0.480}). Only an LLM judge given the rule reads it, and only partly (\rcpt{0.710}, within-pair \rcpt{0.865}).

\subsection{The Crossed Rule-Scenario Benchmark, in Full}
\label{app:crossed}
Construction. Eight domains (ccpa, cybersecurity\_mitre\_attack, data\_act, eu\_ai\_act, finance\_crypto, gdpr, hipaa, policy\_openai) each contribute 25 templates, one per regulatory axis (numeric deadlines, exemption scope, consent basis, jurisdiction, actor type, purpose limitation, and legacy-versus-new-regime conditions), for \rcpt{200} templates and \rcpt{800} rows. Each template supplies a rule with two category-conditioned thresholds and two scenarios that differ only in which category applies, at a shared numeric value strictly between the two thresholds, so the crossed 2$\times$2 design (rule R0/R1 by scenario Sa/Sb) is guaranteed by construction, and which of R0/R1 is the ``natural'' versus ``swapped'' threshold assignment is independently coin-flipped per template to remove any global authoring bias. Because each rule text and each scenario text appears exactly twice in the set, once with each label, rule-only and scenario-only prediction is at chance not merely empirically but by construction, confirmed by the TF-IDF self-check (rule-only, scenario-only, and rule+scenario concatenated all score \rcpt{0.500} AUROC, group-disjoint 5-fold CV by quadruple).

Statistical protocol and quality control. The quadruple is the resampling and reporting unit throughout, never the individual cell, so every rate above (quadruple exact match, rule-flip accuracy, scenario-flip accuracy) is already a proportion over independent quadruples, and we report Wilson 95\% confidence intervals on that proportion directly rather than a Monte Carlo bootstrap over the same units, the two agree in the large-sample limit and the closed form is exact rather than simulated. A predeclared per-domain BH correction (as used for the 20-domain OmniCompliance audit elsewhere in this paper) is not meaningful here, 25 quadruples per domain is too small a cell for a per-domain hypothesis test to be informative, so we report pooled statistics across all 8 domains instead of an underpowered per-domain breakdown. An independent pass manually inspected 45 of the 200 quadruples (180 rows), drawn by an unrelated random seed, for semantic sanity, not label correctness, which is mechanical and not at issue. Forty-one passed cleanly, four had cosmetic issues only, a duplicated article inserted by template substitution in two rule templates and a missing capital letter in two others, all four fixed directly in the released templates, and zero had a genuine construction flaw, no garbled text, no leftover placeholder token, and no scenario stating its own verdict.

Detector matrix, primary model (Llama-3.2-1B-Instruct). Frozen ICS (OmniCompliance-calibrated), \rcpt{0.501} AUROC, \rcpt{0.0}\% quadruple exact match. Fit-on-crossed meandiff, logistic, and MLP probes at the same anchor layer, \rcpt{0.502}, \rcpt{0.507}, and \rcpt{0.501} AUROC, all \rcpt{0.0}\% exact match. Forced-choice judge (Llama-3.1-8B, single-token verdict), \rcpt{0.507} AUROC. Llama Guard 3 and Qwen3Guard, \rcpt{0.500} and \rcpt{0.515} AUROC. LPG-4B, given the rule through its own documented custom-policy channel, \rcpt{0.535} AUROC, \rcpt{1.0}\% exact match. Rachmil et al.'s training-free whitening detector, their own precomputed guard categories and thresholds (github.com/FujitsuResearch/LLM-policy-violation-detection), scored zero-shot with no new calibration on our end, \rcpt{0.498} AUROC, \rcpt{0.0}\% exact match.

Scale replication. Refitting the probe directly on the crossed data recovers a real, scale-dependent signal well short of solved. Llama-3.1-8B logistic reaches \rcpt{0.616} AUROC and \rcpt{7.0}\% exact match (Wilson 95\% CI \rcpt{4.2}--\rcpt{11.4}\%). Qwen3-8B MLP reaches \rcpt{0.670} AUROC and \rcpt{11.0}\% exact match (CI \rcpt{7.4}--\rcpt{16.1}\%), the best cheap-detector result obtained, still a weak effect against \rcpt{6.25}\% chance. Frozen ICS on both larger models stays near chance (\rcpt{0.510} and \rcpt{0.540} AUROC).

Positive control. The identical Llama-3.1-8B judge, given the same rule and scenario but prompted to reason step by step before a final verdict line rather than forced to a single token, is scored on a fixed, seeded \rcpt{40}-quadruple (\rcpt{160}-row) subsample for
compute reasons. It reaches \rcpt{0.849} AUROC, \rcpt{79.5}\%
rule-flip accuracy, \rcpt{78.2}\% scenario-flip accuracy, and \rcpt{74.4}\% quadruple exact match (Wilson 95\% CI \rcpt{58.9}--\rcpt{85.4}\%), confirming the benchmark is solvable by genuine reasoning and the chance-level results above reflect a readout limitation, not a degenerate design.

\subsection{A Worked Example}
\label{app:crossedexample}
The rule-blindness claim above is argued entirely in aggregate numbers. One concrete quadruple, drawn by an independent random seed rather than selected for effect, illustrates what a chance-level detector is actually missing. Domain \texttt{finance\_crypto}, template \texttt{t07}. Rule R0 states a bank may apply simplified due diligence to a dormant retail account below \$1{,}000 and to a dormant institutional account below \$5{,}000. Rule R1 swaps the two limits. Scenario Sa describes a dormant \emph{retail} savings account at Kestrel Bank with a \$2{,}500 balance, and scenario Sb describes the same bank's dormant \emph{institutional} account, also at \$2{,}500. R0$\times$Sa is a violation (\$2{,}500 exceeds the \$1{,}000 retail limit), R0$\times$Sb is compliant (\$2{,}500 is under the \$5{,}000 institutional limit), and R1 reverses both verdicts. Solving all four cells requires reading which account type the scenario names, finding that type's specific threshold inside the rule, and comparing it against the one number the scenario states, exactly the composition step every chance-level detector in the paper's crossed-benchmark table skips.

\subsection{LPG Violated-Policy-Removal Replication}
\label{app:lpgremoval}
LPG's own paper runs a violated-policy-removal probe on two OTHER guards (Qwen3-4B, DynaGuard-8B) and finds low, inconsistent flip-to-safe rates when the violated policy is removed, evidence those guards condition on positional and content priors rather than the supplied clauses. We run the analogous protocol on LPG-4B itself, using \rcpt{200} violating cases from the crossed benchmark, each with its genuine violated rule, a matched permissive-counterpart rule for the identical scenario (the crossed design's own R1 partner), and four synthetic irrelevant decoy policies drawn from other domains, under six conditions, full policy list, remove all policies, remove only the violated policy but keep the decoys, replace the violated policy with its permissive counterpart, shuffle policy order, and add further irrelevant policies. Of \rcpt{182} cases LPG flags unsafe under the full policy list, removing the violated policy entirely, or removing it while leaving the decoys in place, flips the verdict to safe in only \rcpt{44.5\%} of cases either way, so a majority of verdicts do not track whether the violated rule is even present. Substituting the exact permissive counterpart for the violated rule, the sharpest test since the scenario, decoys, and list position are all held fixed and only the substance of the applicable rule changes, flips the verdict in just \rcpt{7.1\%} of cases. Verdict consistency under a pure reordering of the same policy list is \rcpt{92.5\%}, and adding further irrelevant policies alongside an untouched violated rule causes a false flip away from unsafe in only \rcpt{2.7\%} of cases, so LPG is not simply noise-sensitive to list length or order. It also correctly cites the violated rule's own list position in its generated verdict \rcpt{91}--\rcpt{95\%} of the time across the full, shuffled, and irrelevant-added conditions. Read together, LPG reliably points at the right clause when asked which one applies, but the verdict itself barely depends on what that clause actually says, the same dissociation between citation and adjudication the crossed benchmark's chance-level AUROC (the paper's What the Score Measures section) already shows from a different angle.

\begin{table*}[t]
\caption{\textbf{The guards' own turf, seven standard safety benchmarks} (Macro~F1~$\times100$, native decision threshold). The calibrate-once leave-one-distribution-out table is the appropriate cross-domain comparison; threshold-free guard AUROC on these safety sets is deferred to future work. Llama Guard~3~(8B), LPG-4B, and GLiGuard were not cached.}
\label{tab:guards}
\vspace{2mm}
\centering\small
\setlength{\tabcolsep}{9pt}
\fitwidth{%
\begin{tabular}{l rrrrrrr r}
\toprule
Arm & ToxiChat & OpenAIMod & Aegis1.0 & Aegis2.0 & WildGuard & SafeRLHF & BeaverT & Avg \\
\midrule
\multicolumn{9}{l}{\textit{Generative / fine-tuned guard models (Macro F1 $\times100$, native threshold)}} \\
Qwen3Guard-Gen (8B) & 89.1 & 82.5 & 78.2 & 85.7 & 90.0 & 87.1 & 78.0 & 84.4 \\
WildGuard (7B) & 81.3 & 75.6 & 87.7 & 72.1 & 89.9 & 79.1 & 69.1 & 79.3 \\
SIREN & 56.6 & 76.4 & 74.8 & 69.6 & 69.2 & 83.2 & 76.2 & 72.3 \\
HarmBench cls (7B) & 58.5 & 73.0 & 45.9 & 47.7 & 58.0 & 84.4 & 68.9 & 62.3 \\
Llama Guard 3 (1B) & 48.1 & 40.9 & 24.2 & 32.9 & 36.0 & 32.5 & 29.5 & 34.9 \\
\midrule
\multicolumn{9}{l}{\textit{Reference (Macro F1 $\times100$)}} \\
LLM judge (zero-shot, 8B) & 62.0 & 76.8 & 71.2 & 75.5 & 68.1 & 77.0 & 66.2 & 71.0 \\
TF-IDF floor & 75.7 & 75.5 & 71.9 & 75.4 & 74.4 & 81.1 & 68.5 & 74.6 \\
\bottomrule
\end{tabular}
}
\end{table*}

\begin{table*}[t]
\caption{\textbf{Calibration transfer matrix} (ICS AUROC$\uparrow$). Each row's direction fit on that distribution's train, scored on each column's disjoint test; \textbf{bold} = in-domain diagonal. Block means: diagonal \textbf{.849}, within-family .70, cross-family .53. Transfer holds within a family and collapses across it; no single calibration ships everywhere, and the best single calibrator (\texttt{openaimod}, off-diagonal mean .686) still collapses on the opposite family. Calibration $N$ per Table~\ref{tab:splits}.}
\label{tab:transfermat}
\vspace{2mm}
\centering\small
\setlength{\tabcolsep}{9pt}
\fitwidth{%
\begin{tabular}{l cccc c ccccccc}
\toprule
& \multicolumn{4}{c}{\textbf{Compliance} (test)} & & \multicolumn{7}{c}{\textbf{Safety} (test)} \\
\cmidrule(lr){2-5}\cmidrule(lr){7-13}
Calibrated on $\downarrow$ & Omni & Tfin & Tlaw & Tmed & & ToxC & OAI & Aeg1 & Aeg2 & WildG & PKU & Bvr \\
\midrule
\multicolumn{13}{l}{\textit{Compliance-family directions}}\\
OmniCompliance & \textbf{.975} & .471 & .401 & .595 & & .681 & .612 & .696 & .723 & .589 & .630 & .528 \\
TRIDENT-finance & .439 & \textbf{.837} & .689 & .905 & & .452 & .572 & .489 & .544 & .380 & .496 & .525 \\
TRIDENT-law & .598 & .815 & \textbf{.777} & .935 & & .484 & .578 & .466 & .464 & .595 & .572 & .618 \\
TRIDENT-med & .741 & .788 & .676 & \textbf{.916} & & .498 & .594 & .502 & .551 & .525 & .551 & .607 \\
\midrule
\multicolumn{13}{l}{\textit{Safety-family directions}}\\
ToxicChat & .812 & .257 & .384 & .268 & & \textbf{.845} & .723 & .556 & .584 & .668 & .615 & .698 \\
OpenAIMod & .823 & .654 & .458 & .636 & & .826 & \textbf{.898} & .685 & .737 & .779 & .606 & .641 \\
Aegis1 & .875 & .391 & .394 & .510 & & .865 & .769 & \textbf{.833} & .798 & .660 & .729 & .710 \\
Aegis2 & .828 & .360 & .335 & .600 & & .718 & .592 & .700 & \textbf{.780} & .560 & .684 & .727 \\
WildGuard & .622 & .253 & .423 & .446 & & .771 & .701 & .683 & .622 & \textbf{.864} & .834 & .749 \\
PKU-SafeRLHF & .654 & .291 & .346 & .295 & & .759 & .658 & .665 & .694 & .864 & \textbf{.800} & .848 \\
BeaverTails & .560 & .578 & .473 & .766 & & .732 & .640 & .603 & .771 & .830 & .735 & \textbf{.810} \\
\bottomrule
\end{tabular}
}
\end{table*}

\begin{table*}[htbp]
\caption{\textbf{Never-pooled external sets} (AUROC$\uparrow$), pool-calibrated ICS on seven benchmarks sharing no data with any pool. \textbf{bold} = aligned direction; the three row blocks are separate constructs and are not averaged. Prompt-harm sets are read by the safety direction, compliance at chance. Response sets (comply-vs-refuse) are read by the compliance direction along that axis, not regulatory compliance. Unfair-ToS is read above chance for the first time by the shared signal, not the compliance-specialised direction: a compliance direction transfers to unseen scenarios but not to a different compliance genre, and re-fitting on the fine-tuned models of Table~\ref{tab:adapt} still reads these externals at chance (\rcpt{0.44}--\rcpt{0.62}). $^\dagger$ECtHR inconclusive (512-token truncation).}
\label{tab:external}
\vspace{2mm}
\centering\small
\setlength{\tabcolsep}{14pt}
\fitwidth{%
\begin{tabular}{l ccc}
\toprule
External (never-pooled) & \multicolumn{3}{c}{ICS direction calibrated on} \\
\cmidrule(lr){2-4}
 & All-11 Pool & Compliance-Only & Safety-Only \\
\midrule
\multicolumn{4}{l}{\textit{Legal/contractual document labelling} (not rule-conditioned compliance)}\\
Unfair-ToS (LexGLUE) & \textbf{0.612} & 0.515 & 0.688 \\
ECtHR rulings$^\dagger$ & 0.420 & 0.413 & 0.518 \\
\midrule
\multicolumn{4}{l}{\textit{Is this prompt harmful?}}\\
WildJailbreak (prompt) & 0.687 & 0.499 & \textbf{0.767} \\
XSTest (prompt) & 0.855 & 0.502 & \textbf{0.931} \\
JailbreakBench & 0.624 & 0.486 & \textbf{0.649} \\
\midrule
\multicolumn{4}{l}{\textit{Did the response comply or refuse?}}\\
LLM-LAT (response) & 0.996 & 0.997 & 0.980 \\
SORRY-Bench (response) & 0.691 & \textbf{0.751} & 0.536 \\
\bottomrule
\end{tabular}
}
\end{table*}

\noindent As the strictest check, the three pool directions are scored on \emph{seven} never-pooled external benchmarks, two compliance and five safety, sharing no row, benchmark, or collection process with any pool (Table~\ref{tab:external}, all verified disjoint by activation- and text-identity). The pool-calibrated readout fires above chance on never-seen sources and the family structure mostly holds. \textbf{The compliance half of that claim is now tested, and marginally holds}: an external compliance benchmark is read above chance for the first time, but by the \emph{shared} signal, not the compliance-specialised direction, since a compliance direction transfers to unseen \emph{scenarios} but not to a different compliance \emph{genre}, a ceiling that re-fitting on fine-tuned models does not repair.

\begin{table*}[t]
\caption{\textbf{The efficiency axis, stated conditionally.} ICS's zero-extra-pass property holds only when the scored text already passed through the monitored model with the anchor-layer activation retained; scoring external text costs one pass, like the other internal-representation probes. Only the generative guards need no hidden-state access at all. The \rcpt{$\sim$0} figure is the readout alone (one dot product), excluding hook overhead, retention memory, and transfer, none of which were measured; treat it as a lower bound. ms/case measured in E-GUARDBENCH. GLiGuard did not run (n/a); $\sim$-marked example counts are order-of-magnitude public estimates.}
\label{tab:efficiency}
\vspace{2mm}
\centering\small
\setlength{\tabcolsep}{12pt}
\fitwidth{%
\begin{tabular}{lrrccr}
\toprule
 & Trainable & Labelled & \multicolumn{2}{c}{Extra pass over the monitored model} & Readout \\
\cmidrule(lr){4-5}
Arm & Params & Examples & Activations Retained & External Text & ms/case \\
\midrule
\textbf{ICS} (ours) & \textbf{0} & \textbf{10 pairs/dist.} & \textbf{0} & 1 & \textbf{$\sim$0} \\
GradSafe & \textbf{0} & 16 pairs & 1 fwd $+$ bwd & 1 fwd $+$ bwd & 200 \\
SIREN & 14M & $\sim$$10^5$ & 0 & 1 & 8.4 \\
GLiGuard & 300M & $\sim$$10^5$ & 0 & 1 & n/a \\
Llama Guard 3 (8B) & 8B & $10^4$--$10^5$ & 1 autoregressive & 1 autoregressive & 31.9 \\
WildGuard (7B) & 7B & $\sim$87K & 1 autoregressive & 1 autoregressive & 31.8 \\
Qwen3Guard-Gen (8B) & 8B & $\sim$$10^6$ & 1 autoregressive & 1 autoregressive & 45.3 \\
\bottomrule
\end{tabular}
}
\end{table*}

\begin{figure*}[htbp]\centering
\includegraphics[width=\columnwidth]{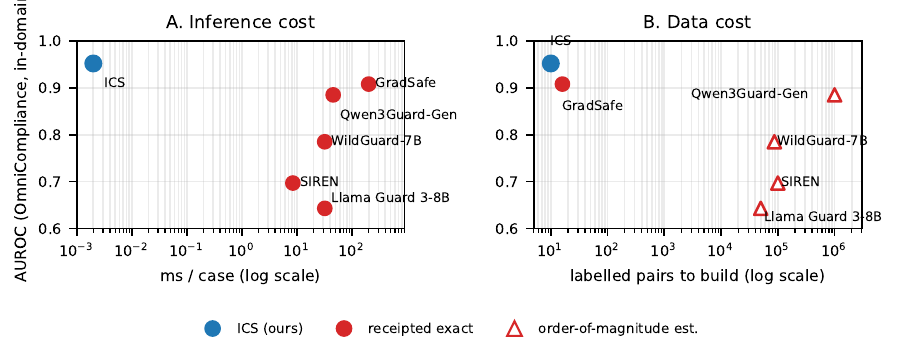}
\caption{\textbf{Accuracy and estimated scoring cost.} \emph{Left}, inference cost (ms/case); \emph{right}, data cost (labelled pairs, log scale), both vs.\ AUROC on outcome-ablated OmniCompliance. Open triangles mark public estimates not receipted; filled points are exact (Table~\ref{tab:efficiency}). ICS gives competitive accuracy at its cost, beaten only by GradSafe at 200ms and by nothing at its own; a trained probe on the same activations is slightly more accurate still (the paper's in-domain detector-comparison table, the paper's Scope and Limitations section), so the advantage plotted here is cost, not a ceiling on accuracy. Cost estimates exclude activation-hook, retention, and transfer overhead (Table~\ref{tab:efficiency}).}
\label{fig:pareto}
\end{figure*}
\clearpage
\FloatBarrier
\section{Full Per-Domain Gate Table}
\label{app:gate}
Table~\ref{tab:gatefull} gives all 20 domains of the pre-registered gate; the main text shows 8. \textsc{foundation\_rights} is the one domain that both fails this gate at $n{=}100$ against a mid-range floor and moves adversely under the rule-disjoint restriction (\rcpt{0.909}$\to$\rcpt{0.731}, \S\ref{app:s3}).

\begin{table*}[htbp]
\caption{The pre-registered gate, all 20 domains. Every passing domain is $n{=}100$; failures are heterogeneous: small-$n$, ceiling-saturated, or the corrected sentiment floor closing to within 0.002--0.012.}
\label{tab:gatefull}
\vspace{2mm}
\centering\small
\setlength{\tabcolsep}{9pt}
\fitwidth{%
\begin{tabular}{lrrrlrc}
\toprule
Domain & $n$ & \textbf{ICS} & Max.\ trivial floor & (Which floor) & $\Delta$ & Gate \\
\midrule
Fin.\ cross-border & 63 & 0.961 & 0.960 & \textsc{sst2} & $+0.002$ & -- \\
Data Act & 100 & 0.964 & 0.894 & \textsc{sst2} & $+0.070$ & $\checkmark$ \\
Policy: Google & 100 & 0.970 & 0.919 & \textsc{sst2} & $+0.051$ & $\checkmark$ \\
SB-35 & 100 & 0.935 & 0.885 & \textsc{sst2} & $+0.050$ & $\checkmark$ \\
Policy: OpenAI & 100 & 0.926 & 0.864 & \textsc{sst2} & $+0.063$ & $\checkmark$ \\
Fin.\ AML/CTF & 100 & 0.965 & 0.918 & \textsc{sst2} & $+0.047$ & $\checkmark$ \\
HIPAA & 100 & 0.983 & 0.941 & \textsc{sst2} & $+0.042$ & $\checkmark$ \\
Policy: GitHub & 100 & 0.909 & 0.868 & \textsc{sst2} & $+0.042$ & $\checkmark$ \\
Policy: Reddit & 100 & 0.921 & 0.861 & \textsc{sst2} & $+0.060$ & $\checkmark$ \\
Fin.\ e-money & 100 & 0.995 & 0.970 & \textsc{sst2} & $+0.025$ & $\checkmark$ \\
EU AI Act & 100 & 0.986 & 0.938 & \textsc{sst2} & $+0.048$ & $\checkmark$ \\
Fin.\ crypto & 100 & 0.997 & 0.985 & \textsc{sst2} & $+0.012$ & -- \\
CCPA & 100 & 0.986 & 0.946 & \textsc{tfidf} & $+0.040$ & $\checkmark$ \\
GDPR & 100 & 0.863 & 0.862 & \textsc{sst2} & $+0.001$ & -- \\
Edu: online & 72 & 0.997 & 0.999 & \textsc{sst2} & $-0.003$ & -- \\
Policy: X & 100 & 0.936 & 0.928 & \textsc{tfidf} & $+0.008$ & -- \\
Cybersec. & 36 & 0.997 & 0.989 & \textsc{tfidf} & $+0.008$ & -- \\
Found.\ rights & 100 & 0.909 & 0.902 & \textsc{tfidf} & $+0.006$ & -- \\
Edu: discrim. & 16 & 0.875 & 0.887 & \textsc{tfidf} & $-0.012$ & -- \\
Edu: integrity & 36 & 0.963 & 0.993 & \textsc{tfidf} & $-0.030$ & -- \\
\midrule
\textit{Mean} & & \textit{0.952} & \textit{0.925} & & \textit{+0.027} & \textbf{11/20} \\
\bottomrule
\end{tabular}
}
\end{table*}


\FloatBarrier
\section{Lexical Floors Across Benchmarks}
\label{sec:degeneracy}

\noindent\textbf{Four of seven public compliance benchmarks cannot adjudicate a probe.} If the label is a property of the scenario rather than of the rule--scenario relation (the paper's What the Score Measures section), it should be recoverable without a model at all. It is. The verdict-narration diagnosis of the paper's Evaluation Validity section is not one dataset's quirk: of seven public benchmarks tested, four are lexically degenerate, a fifth is borderline, and only two support measurement at all. On DynaBench \citep{hoover2025dynaguard}, the test set is lexically shortcut-able: a policy-blind bag of words over the transcript alone reaches within-test AUROC \rcpt{0.982} and F1 \rcpt{0.88} under cross-validation. We verified this is a property of the test set's construction, \emph{not} a same-split match to their calibrated result: fit on the benchmark's own published training split and evaluated on their test rows, the same bag of words is near chance (F1 \rcpt{0.51}--\rcpt{0.61}, verification code released with the evaluation protocol), so the shortcut bounds the benchmark, and does not beat their training-free method head-to-head. ICS numbers on DynaBench are truncation-limited at 512 tokens; the floor-exceeds-probe ordering survives a truncation control. Table~\ref{tab:benchmarks}'s ICS figure for DynaBench (\rcpt{0.784}) is measured under this gauntlet's own uniform within-test protocol (same 5-fold cross-validation as every other benchmark here, for comparability across the fourteen-benchmark table); the stricter head-to-head against Rachmil et al.'s own published train/test split \citep{rachmil2025whitening} is reported separately in the Related Work comparison against that method (\S\ref{subsec:related-work}). LPG-4B, run on the identical 543-row DynaBench test split, extends this to a policy-reasoning guard rather than an activation probe. Given the correct policy it reaches AUROC \rcpt{0.681}, given no policy at all AUROC \rcpt{0.913}, and given a policy for an unrelated case AUROC \rcpt{0.687}, so the correct policy performs no better than a wrong one and both perform far worse than no policy at all, the reverse of what a rule-conditioned guard should show. This tracks the same degeneracy diagnosis, a transcript-only bag-of-words already reaches \rcpt{0.982} on this test set, so the policy block mostly displaces useful transcript content inside LPG's \rcpt{1024}-token window rather than adding signal. On SafePyramid, TF-IDF over the \emph{rule text alone}, never reading the conversation, reaches \rcpt{0.90+} at every difficulty level, while the conversation alone (grouped CV) scores exactly 0.500: the label is predictable from which rule was asked about; on a surface-matched flip-pair core (identical rule text, opposite label), every lexical feature collapses to \rcpt{0.5000}. On AIReg-Bench, the one benchmark with human-expert labels, TF-IDF over the document alone reaches \rcpt{0.957} (grouped 5-fold \rcpt{0.975}): expert annotation does not repair outcome-first construction, because the shortcut is the document's stylistic register, not annotator noise. CompliBench is borderline (\rcpt{0.879}), and we additionally found its per-turn guideline metadata leaks the label outright (\rcpt{0.990}; 823 of 830 violating turns carry the modified, violation-inducing directive); we report fair reconstructed floors instead. The pattern is not specific to our seven: on the third-party \textsc{trident} benchmark \citep{trident2025} a policy-blind TF-IDF scores \rcpt{0.999} on the benchmark's own texts (a request-vs-advisory form shortcut) and \rcpt{0.910} on a cleaner response-based contrast, beating ICS (\rcpt{0.802}) on 2 of 3 domains; ICS does not clear the floor there either. Prior work reports no lexical floor on any of these.
\begin{table*}[htbp]
\caption{\textbf{Fourteen benchmarks under one gauntlet.} ``Floor'' is the strongest \emph{policy-blind} lexical model (a bag of words that never reads the rule). A benchmark is \emph{degenerate} when the floor is high \emph{and} the probe does not significantly clear it. Compliance floors run 0.57--0.98 and 4/7 are degenerate (including the human-expert-labelled one); safety floors 0.77--0.90, 2/7 degenerate. \textbf{The diagnosis is specific to compliance benchmarks, not to activation probing:} where labels come from a mechanical verifier the floor collapses to 0.574. Safety-suite nulls are the ICS layer-selection budget (never 0.5).}
\label{tab:benchmarks}
\vspace{2mm}
\centering\small
\setlength{\tabcolsep}{3pt}
\fitwidth{%
\begin{tabular}{llrrlrrrl}
\toprule
Benchmark & Labels from & $n$ & Base rate & Scoring unit & Lex.\ floor & \textbf{ICS (ours)} & Budget null & Verdict \\
\midrule
\multicolumn{9}{l}{\textit{Compliance: does this scenario violate this rule?}} \\
OmniCompliance & enforcement text & 16--100 & .500 & rule--scenario pair & 0.896 & 0.952 & 0.714 & degenerate \\
DynaBench & narrated verdict & 543 & .492 & transcript & \textbf{0.982} & 0.784 & 0.576 & degenerate \\
SafePyramid & rule text \emph{alone} & 77{,}755 & .344 & conversation--rule & 0.90+ & 0.871 & 0.673 & degenerate \\
AIReg-Bench & \textbf{human experts} & 120 & .367 & documentation & \textbf{0.957} & 0.943 & 0.708 & degenerate \\
CompliBench & narrated $+$ metadata & 2{,}713 & .306 & assistant turn & 0.879 & 0.700 & 0.549 & borderline \\
FlexBench & moderation labels & 2{,}278 & .375 & response & 0.802 & 0.832 & 0.603 & \textbf{clean} \\
IFEval & \textbf{mechanical verifier} & 541 & .481 & prompt & \textbf{0.574} & 0.628 & 0.530 & \textbf{clean} \\
\midrule
\multicolumn{9}{l}{\textit{Safety: is this content harmful? (no rule field to ablate)}} \\
ToxicChat & human annotation & 9{,}743 & .072 & prompt & \textbf{0.903} & 0.884 & 0.576 & degenerate \\
PKU-SafeRLHF & preference labels & 10{,}000 & .519 & prompt$+$response & \textbf{0.895} & 0.903 & 0.580 & degenerate \\
OpenAI Moderation & taxonomy & 1{,}664 & .309 & prompt & 0.850 & 0.893 & 0.560 & clean \\
Aegis 2.0 & human annotation & 10{,}000 & .509 & prompt & 0.836 & 0.809 & 0.588 & clean \\
WildGuard & human annotation & 1{,}725 & .437 & prompt & 0.826 & 0.918 & 0.612 & clean \\
Aegis 1.0 & human annotation & 10{,}000 & .681 & prompt & 0.811 & 0.752 & 0.538 & clean \\
BeaverTails & human annotation & 10{,}000 & .582 & prompt$+$response & 0.766 & 0.782 & 0.574 & clean \\
\midrule
\textit{Compliance} & & & & & \textit{0.57--0.98} & & & \textbf{4/7 degen.} \\
\textit{Safety} & & & & & \textit{0.77--0.90} & & & \textbf{2/7 degen.} \\
\bottomrule
\end{tabular}
}
\end{table*}

Two benchmarks are clean, and they adjudicate against the probe. On IFEval (TF-IDF \rcpt{0.574}; labels from a mechanical verifier, not narrated verdicts), ICS beats the trivial form-counting floor in \rcpt{0 of 7} categories, in every configuration tested. On FlexBench (TF-IDF \rcpt{0.68--0.80}), ICS shows a weak, CI-overlapping lead on the response track at strict severity (\rcpt{0.832 vs 0.802}), inverts at loose severity, and on the prompt track the zero-supervision refusal direction matches supervised TF-IDF outright. Where the floor is honest but high (DynaBench, SafePyramid, CompliBench), it \emph{exceeds} the probe, the inverse of the OmniCompliance ordering. The pattern across seven benchmarks: the probe's apparent advantage is largest exactly where the benchmark is least able to adjudicate it.

\subsection{Guard Models on the Same External and Degeneracy Benchmarks}
\label{sec:guardgap}
Tables~\ref{tab:external} and~\ref{tab:benchmarks} report ICS on thirteen benchmarks that, until this pass, had no deployed-guard comparison. We close that gap: the same five guards benchmarked elsewhere in this paper (Llama Guard~3 (1B), WildGuard, HarmBench-cls, Qwen3Guard-Gen (8B), and an 8B zero-shot LLM judge), scored zero-shot (native mode, no threshold fitting) on the identical texts and labels (Tables~\ref{tab:guardgapexternal} and~\ref{tab:guardgapdegeneracy}). Benchmarks whose larger class exceeded \rcpt{1{,}500} examples were stratified-capped at \rcpt{1{,}500} per class (seed 42) for tractability; capped benchmarks and the number of rows dropped: wildjailbreak (\rcpt{500}), llmlat (\rcpt{6{,}895}), sorry\_bench (\rcpt{332}), unfair\_tos (\rcpt{6{,}716}), ecthr\_a (\rcpt{8{,}257}), safepyramid (\rcpt{21{,}000}), complibench (\rcpt{383}), flexbench (\rcpt{278}); jbb, xstest, dynabench, aireg, and ifeval were scored in full.

\begin{table*}[t]
\caption{\textbf{Guard models on the never-pooled external OOD set} (AUROC$\uparrow$), closing the gap in Table~\ref{tab:external}. \textbf{bold} = best guard per row. On the two response-harm sets, HarmBench-cls (scores the full request+response against a harm behaviour) reads correctly (\rcpt{$>0.999$}, \rcpt{0.775}); WildGuard and the LLM-judge score \emph{below chance} (\rcpt{0.331}/\rcpt{0.255}, \rcpt{0.129}/\rcpt{0.397}), verified not a label bug, since their native templates ask about a prompt or "the request," not whether a response fulfilled it, the same prompt-vs-response sensitivity as the paper's Comparison with Deployed Guards section. $^\dagger$ECtHR inconclusive (as Table~\ref{tab:external}).}
\label{tab:guardgapexternal}
\vspace{2mm}
\centering\small
\setlength{\tabcolsep}{3pt}
\fitwidth{%
\begin{tabular}{lr rrrrr}
\toprule
External (never-pooled) & ICS (all-11 pool) & Llama Guard~3 (1B) & WildGuard & HarmBench-cls & Qwen3Guard-Gen & LLM-judge (8B) \\
\midrule
Unfair-ToS (LexGLUE) & 0.612 & 0.619 & 0.626 & 0.517 & 0.333 & 0.489 \\
ECtHR rulings$^\dagger$ & 0.420 & 0.479 & 0.487 & 0.485 & 0.526 & 0.419 \\
WildJailbreak (prompt) & 0.687 & 0.799 & \textbf{0.972} & 0.646 & 0.945 & 0.864 \\
XSTest (prompt) & 0.855 & 0.917 & \textbf{0.990} & 0.858 & 0.973 & 0.843 \\
LLM-LAT (response) & 0.996 & \textbf{$>0.999$} & 0.331 & $>0.999$ & 0.916 & 0.129 \\
SORRY-Bench (response) & 0.691 & 0.671 & 0.255 & \textbf{0.775} & 0.320 & 0.397 \\
JailbreakBench & 0.624 & \textbf{0.941} & 0.966 & 0.835 & 0.942 & 0.785 \\
\bottomrule
\end{tabular}
}
\end{table*}

\begin{table*}[t]
\caption{\textbf{Guard models on the six degeneracy-audit benchmarks without a prior guard comparison} (AUROC$\uparrow$), closing the gap in Table~\ref{tab:benchmarks}. \textbf{bold} = best guard per row. Every guard sits near chance on DynaBench, SafePyramid, and CompliBench -- the same three domains where \S\ref{sec:degeneracy} finds ICS's own apparent edge is largest exactly where the honest floor is high, so this is not a case of guards underperforming ICS by an unfair comparison: no detector in this class, first-party or deployed, reads these three domains well.}
\label{tab:guardgapdegeneracy}
\vspace{2mm}
\centering\small
\setlength{\tabcolsep}{6pt}
\fitwidth{%
\begin{tabular}{lr rrrrr}
\toprule
Benchmark & ICS (Table~\ref{tab:benchmarks}) & Llama Guard~3 (1B) & WildGuard & HarmBench-cls & Qwen3Guard-Gen & LLM-judge (8B) \\
\midrule
DynaBench & 0.784 & 0.571 & 0.554 & 0.574 & 0.527 & 0.516 \\
SafePyramid & 0.871 & 0.506 & 0.522 & 0.509 & 0.503 & 0.491 \\
AIReg-Bench & 0.943 & \textbf{0.603} & 0.555 & 0.445 & 0.445 & 0.179 \\
CompliBench & 0.700 & 0.518 & \textbf{0.522} & 0.491 & 0.506 & 0.428 \\
FlexBench & 0.832 & \textbf{0.849} & 0.742 & 0.814 & 0.800 & 0.677 \\
IFEval & 0.628 & \textbf{0.559} & 0.560 & 0.549 & 0.536 & 0.522 \\
\bottomrule
\end{tabular}
}
\end{table*}

\paragraph{The LLM judge splits along task type, not benchmark family.} A pattern spans both tables that neither caption states alone: the 8B zero-shot judge is the strongest or a competitive guard on every classic prompt-harm set (\rcpt{0.864} wildjailbreak, \rcpt{0.785} jbb, \rcpt{0.843} xstest) but sits at or below chance on nearly every regulatory or legal-compliance benchmark (\rcpt{0.179} aireg, \rcpt{0.428} complibench, \rcpt{0.419} ecthr\_a, \rcpt{0.489} unfair\_tos, \rcpt{0.397} sorry\_bench). The judge is prompted with a single fixed question -- \emph{is the request above harmful or unsafe} -- which is the right question for a jailbreak benchmark and the wrong one for a rule-adherence benchmark: nothing in that prompt asks whether a specific rule was satisfied. This is not a new, unrelated failure mode; it is the paper's central rule-blindness finding (the paper's What the Score Measures section) appearing again in a different guise, on benchmarks outside the pool used to establish it.

\paragraph{The selection floor is not a compliance quirk, and it is not outgrown.} The same budget-matched null inflates random directions on unrelated probing settings (Azaria--Mitchell true/false and SST-2) to \rcpt{0.50--0.65} against no-selection controls at \rcpt{0.49--0.51}, so the correction applies to layer-selected probes generally. Two qualifications: the magnitude is setting-specific (compliance's \rcpt{0.71} is the highest we measured; one topic is essentially uninflated), so a floor must be measured against the arm's own search, not transferred; and it does not decay with data, since a predicted $c/\sqrt n$ fit is refuted ($r^2=\rcpt{-23}$; a plateau fits at $r^2=\rcpt{0.96}$, floor \rcpt{0.09}), with SST-2 still at \rcpt{0.586} at \rcpt{2000} pairs. Details and the transfer table are in this section.

\begin{figure}[htbp]\centering
\resizebox{0.4\columnwidth}{!}{%
\begin{tikzpicture}[font=\small, >=Latex,
cal/.style={draw, rounded corners, fill=blue!8, align=center, minimum height=9mm, minimum width=34mm, inner sep=3pt},
tst/.style={draw, rounded corners, fill=green!14, align=center, minimum height=9mm, minimum width=34mm, inner sep=3pt},
dir/.style={draw, rounded corners, fill=red!8, align=center, minimum height=9mm, minimum width=34mm, inner sep=3pt},
arr/.style={-{Latex[length=2mm]}, semithick},
ttl/.style={font=\footnotesize\itshape, align=center, text width=40mm}]
\hyphenpenalty=10000 \exhyphenpenalty=10000
\node[ttl] (titleA) {\textbf{In-domain}: calibrate and test on the same distribution};
\node[cal, below=6mm of titleA] (ca) {calibration pairs\\from $D$};
\node[dir, below=10mm of ca] (da) {ICS\\direction};
\node[tst, below=10mm of da] (ta) {test:\\held-out $D$};
\draw[arr] (ca)--(da); \draw[arr] (da)--(ta);
\node[ttl, right=20mm of titleA] (titleB) {\textbf{Calibrate-once}: $D$ withheld from calibration, met only at test};
\node[cal, below=6mm of titleB] (cb) {calibration pairs\\from $A,B,C,\dots\ (\neq D)$};
\node[dir, below=10mm of cb] (db) {one ICS\\direction};
\node[tst, below=10mm of db] (tb) {test: $D$\\(\emph{unseen})};
\draw[arr] (cb)--(db); \draw[arr] (db)--(tb);
\end{tikzpicture}}
\caption{\textbf{Calibration vs.\ test data, and the calibrate-once guard test.} ICS is built from a small labelled \emph{calibration} set (pairs give the direction; a validation slice fixes layer and threshold) and scored on a \emph{disjoint} test set. \emph{Left:} per-distribution, with calibration and test from the same $D$. \emph{Right:} calibrate-once, where $D$ is withheld from calibration and appears only at test, so the direction meets it unseen, as a deployed guard meets a new input (the paper's calibrate-once leave-one-distribution-out table).}
\label{fig:calibtest}
\end{figure}
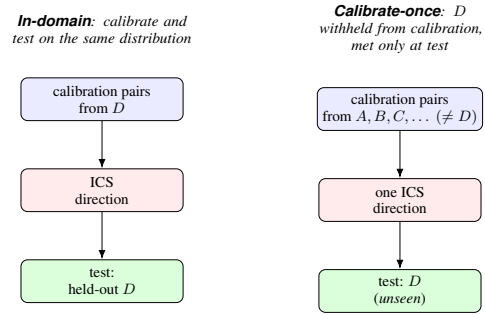

\begin{figure}[htbp]
\centering
\includegraphics[width=0.5\columnwidth]{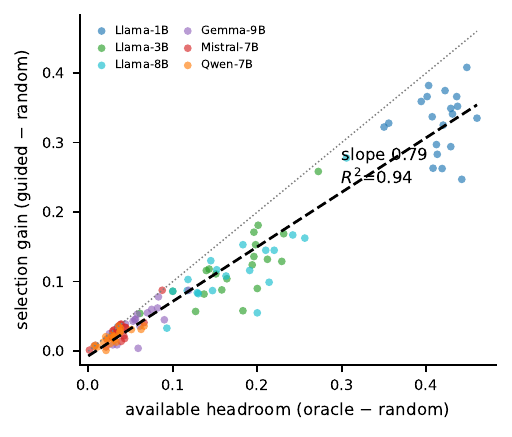}
\caption{\textbf{Selection recovers a near-constant fraction of the available headroom, on every model.} Each point is one (model, domain); $x$ is what the oracle could win over random, $y$ is what ICS-guided selection actually wins. The dashed fit has slope \rcpt{0.79} through the origin ($R^2=\rcpt{0.94}$, 120 points); the dotted line is the oracle ($y{=}x$). Larger models cluster near the origin because their room is small, not because selection weakens; they sit on the same line.}
\label{fig:headroom}
\end{figure}

\section{Split Leakage Audit}
\label{app:s3}
Splits are case-level; \rcpt{68\%} of test cases share a \texttt{source\_rule} with the train slice. Restricting test to rule-disjoint cases leaves AUROC unchanged (\rcpt{0.962} vs.\ \rcpt{0.956}, restricted $\ge$ full in 13/19 domains), so rule overlap does not inflate the headline. One adverse domain (foundational rights, \rcpt{$0.909 \to 0.731$}); power is weakest exactly where leakage is heaviest. The rule-disjoint restriction is computable in 19/20 domains; only \texttt{foundation\_rights} moves adversely (\rcpt{0.909} to \rcpt{0.731}). This domain's rule overlap (\rcpt{84.5\%} of its test cases) is above the dataset mean, so restriction leaves only \rcpt{31} residual cases (\rcpt{18} violating, \rcpt{13} compliant) against \rcpt{200} in the full test set -- the smallest surviving residual of any domain whose full-set ICS score is not already near ceiling. Eight other domains have a comparably small residual ($n\le30$); all eight have a full-set ICS score of \rcpt{0.92} or above and land at a restricted AUROC of exactly \rcpt{1.0} (a ninth, \textsc{edu\_discrimination\_us\_edu\_dept}, restricts to a single case and is not computable at all -- the source of this section's ``computable in 19/20 domains''). \textsc{foundation\_rights}, whose full-set score (\rcpt{0.909}) is the third-lowest of all 20 domains (after \textsc{gdpr}'s \rcpt{0.863} and \textsc{edu\_discrimination\_us\_edu\_dept}'s \rcpt{0.875}, the latter itself one of the small-$n$ gate failures, not this section's concern), is the first domain among these small-residual cases where the signal is weak enough at full scale to show through rather than round up to ceiling. \textsc{gdpr} has a lower full-set score still (\rcpt{0.863}) but a far lower rule-overlap fraction (\rcpt{51.5\%}), so its residual stays large (\rcpt{97} cases) and its restricted AUROC is stable (\rcpt{0.863}$\to$\rcpt{0.880}). The adverse swing is therefore a small-$n$ consequence of an already-weak full-set signal meeting unusually high rule overlap, not evidence of a distinct failure mode; it does not explain why the full-set signal is weak on this domain in the first place (the paper's Scope and Limitations section).



\FloatBarrier
\section{Cross-Family Generalisation (Leave-One-Family-Out)}
\label{app:family}

\begin{table*}[htbp]
\caption{\textbf{Cross-validation splits.} Per-distribution train/val/test counts (balanced 50/50, seed 42, group-disjoint) pooled for the calibrate-once (the paper's calibrate-once leave-one-distribution-out table) and transfer-matrix (Table~\ref{tab:transfermat}) experiments.}
\label{tab:splits}
\vspace{2mm}
\centering\small
\setlength{\tabcolsep}{20pt}
\fitwidth{%
\begin{tabular}{llrrr}
\toprule
Distribution & Family & Train & Validation & Test \\
\midrule
OmniCompliance & compliance & 320 & 320 & 160 \\
\textsc{trident}-finance & compliance & 240 & 240 & 122 \\
\textsc{trident}-law & compliance & 210 & 208 & 110 \\
\textsc{trident}-med & compliance & 160 & 160 & 84 \\
\midrule
ToxicChat & safety & 320 & 320 & 160 \\
OpenAIMod & safety & 320 & 320 & 160 \\
Aegis1 & safety & 320 & 320 & 157 \\
Aegis2 & safety & 320 & 320 & 160 \\
WildGuard & safety & 320 & 320 & 160 \\
PKU-SafeRLHF & safety & 320 & 320 & 160 \\
BeaverTails & safety & 320 & 320 & 160 \\
\bottomrule
\end{tabular}
}
\end{table*}

The main leave-one-distribution-out test (the paper's Generalisation Beyond the Calibration Distribution section) holds out one distribution but keeps its family siblings in the calibration pool (splits, Table~\ref{tab:splits}). The stricter test withholds an \emph{entire} family. \textbf{Compliance$\to$Safety} (fit on the four compliance distributions only, test on the seven safety benchmarks): mean AUROC \rcpt{0.552}, every target in 0.50--0.60. \textbf{Safety$\to$Compliance} (fit on the seven safety distributions only, test on OmniCompliance $+$ \textsc{trident}): mean \rcpt{0.476}: OmniCompliance survives (\rcpt{0.784}, its generic violation signal is broad) but the three \textsc{trident} domains invert below chance (finance \rcpt{0.294}, law \rcpt{0.314}). In-domain ceilings stay high (0.78--0.98), so the collapse is purely the cross-family tax, not weak targets. Zero target-family test rows appear in the train-family pool (asserted). The calibrate-once direction is therefore domain-type-specific: it must be calibrated on the family it will monitor.

\FloatBarrier
\section{Applying Guards on Input vs.\ Response}
\label{app:guardio}
\begin{table*}[htbp]
\caption{\textbf{Input vs.\ response is a real knob, and response wins.} Each generative guard applied to the case as an \emph{input} to screen vs.\ a \emph{response} to judge (\textbf{bold} = better mode); response helps almost everywhere, decisively for WildGuard ($+0.32$ mean AUROC, near-chance on \textsc{trident} input). $^\dagger$HarmBench is natively a response classifier. ICS has no input/response slot, reading the case activation directly. This table is the reference for guard mode; the headline tables now consistently credit each guard its stronger, response mode recorded here.}
\label{tab:guardio}
\vspace{2mm}
\centering\small
\setlength{\tabcolsep}{9pt}
\fitwidth{%
\begin{tabular}{ll ccccc cc}
\toprule
Guard & Mode & OmniCompliance AUROC & AUPRC & F1 (native) & F1 (recal.) & \textsc{trident}-fin & \textsc{trident}-law & \textsc{trident}-med \\
\midrule
Llama Guard 3 (1B) & input & 0.689 & 0.698 & 0.312 & 0.619 & 0.807 & 0.741 & \textbf{0.903} \\
 & response & \textbf{0.749} & \textbf{0.758} & 0.245 & \textbf{0.681} & \textbf{0.837} & \textbf{0.778} & 0.896 \\
WildGuard (7B) & input & 0.785 & 0.787 & \textbf{0.113} & 0.702 & 0.512 & 0.567 & 0.599 \\
 & response & \textbf{0.875} & \textbf{0.886} & 0.128 & \textbf{0.814} & \textbf{0.874} & \textbf{0.781} & \textbf{0.967} \\
HarmBench cls (7B)$^\dagger$ & input & 0.720 & 0.724 & 0.072 & 0.658 & 0.801 & \textbf{0.808} & 0.890 \\
 & response & \textbf{0.735} & \textbf{0.738} & \textbf{0.105} & \textbf{0.673} & \textbf{0.812} & \textbf{0.808} & \textbf{0.894} \\
Qwen3Guard-Gen (8B) & input & 0.885 & 0.884 & 0.146 & 0.807 & 0.776 & 0.674 & 0.821 \\
 & response & \textbf{0.908} & \textbf{0.905} & \textbf{0.191} & \textbf{0.833} & \textbf{0.854} & \textbf{0.774} & \textbf{0.946} \\
\bottomrule
\end{tabular}%
} 
\end{table*}

A compliance case is a Rule plus a Scenario. A generative moderation guard can read it as an \emph{input} to screen (Scenario in the user slot) or as a \emph{response} to judge (Scenario in the assistant slot, Rule as policy); the two are distinct tasks for a guard trained on live prompts or outputs. We run each generative guard (Llama Guard~3, WildGuard, HarmBench, Qwen3Guard) in both modes on OmniCompliance and the three \textsc{trident} domains, scoring each mode's AUROC/AUPRC and native/recalibrated F1 separately on the same seed-42 test splits; the head-to-head in the paper's Generalisation Beyond the Calibration Distribution section credits each guard its stronger mode. Table~\ref{tab:guardio} reports the per-benchmark AUROC in both modes. Response mode is the stronger role for all four guards on compliance, and the gap is decisive for WildGuard: its prompt-slot head is near-chance on \textsc{trident} (0.51--0.60) and only the response head recovers ($+0.32$ AUROC on average). This is exactly why a single fixed mapping (as in the paper's Generalisation Beyond the Calibration Distribution section's harness, which sends WildGuard to the prompt slot) can understate a guard, because it may be reading its weaker role. We therefore report each guard's stronger mode.

\begin{table*}[htbp]
\caption{\textbf{All metrics on outcome-ablated OmniCompliance} (mean over 20 domains). At native harm-tuned threshold guards rarely fire on compliance (recall 0.08--0.22); recalibrating recovers most of it (WildGuard $0.11\to0.70$, Qwen3Guard $0.15\to0.81$), still below ICS's val-calibrated F1 of 0.904 (a different protocol from the paper's in-domain detector-comparison table's ten-pair-budget 0.897, not a second measurement). ICS is best-or-tied on 5 of 8 metrics (AUROC, F1, Accuracy, Recall, FPR@95) at zero parameters, though the MLP probe on the same activations edges it out; LPG-4B leads on AUPRC (0.947) and Precision (0.923), and Qwen3Guard-Gen on Specificity (0.998). SIREN's params count its \rcpt{22}MB probe head only, riding the same frozen backbone ICS does. Guard AUROCs credit each guard its stronger, response mode throughout (Table~\ref{tab:guardio}), recomputed and verified per-domain-then-mean from the same raw scores, released with the evaluation protocol.}
\label{tab:metrics}
\vspace{2mm}
\centering\small
\setlength{\tabcolsep}{7pt}
\fitwidth{%
\begin{tabular}{llrrrrrrrr}
\toprule
Arm & Params & AUROC & AUPRC & F1 & Acc & Prec & Rec & FPR@95 & Spec \\
\midrule
\textbf{ICS} (ours) & 0 / 10 pairs & \textbf{0.952} & \textbf{0.940} & \textbf{0.904} & \textbf{0.903} & 0.895 & \textbf{0.915} & \textbf{0.164} & 0.891 \\
SIREN & 14M & 0.697 & 0.692 & 0.282 & 0.560 & 0.651 & 0.221 & 0.778 & 0.900 \\
GLiGuard & 300M & 0.703 & 0.692 & 0.292 & 0.563 & 0.740 & 0.190 & 0.785 & 0.935 \\
MLP probe (same activations) & $\sim$0.1M & 0.966 & 0.963 & 0.916 & 0.916 & 0.917 & 0.915 & 0.133 & 0.917 \\
Llama Guard 3 (1B) & 1B & 0.749 & 0.697 & 0.318 & 0.577 & 0.830 & 0.205 & 0.860 & 0.949 \\
Llama Guard 3 (8B) & 8B & 0.643 & 0.680 & 0.200 & 0.553 & 0.813 & 0.122 & 0.903 & 0.984 \\
Qwen3Guard-Gen (8B) & 8B & 0.908 & 0.885 & 0.145 & 0.540 & 0.888 & 0.081 & 0.482 & 0.998 \\
WildGuard (7B) & 7B & 0.875 & 0.787 & 0.113 & 0.527 & 0.874 & 0.064 & 0.655 & 0.990 \\
HarmBench cls (7B) & 7B & 0.735 & 0.738 & 0.105 & 0.530 & 0.837 & 0.063 & 0.763 & 0.997 \\
Latent Policy Guard (4B) & 4B & 0.943 & 0.947 & 0.847 & 0.861 & \textbf{0.923} & 0.785 & 0.242 & 0.936 \\
LLM judge (zero-shot, 8B) & 8B & 0.912 & 0.918 & 0.840 & 0.830 & 0.807 & 0.886 & 0.374 & 0.775 \\
TF-IDF floor & --- & 0.896 & 0.900 & 0.822 & 0.821 & 0.821 & 0.826 & 0.426 & 0.815 \\
\bottomrule
\end{tabular}
}
\end{table*}

\FloatBarrier
\section{Matched-Threshold Calibration and Cross-Model Adaptation}
\label{app:matchedcalib}

\subsection{Training-free comparators and model-specific recalibration, in full}
The training-free story generalises, and ICS is cheapest. The result is not idiosyncratic to difference-of-means: GradSafe \citep{xie2024gradsafe}, a gradient-based training-free detector, also beats every deployed guard on compliance (AUROC \rcpt{0.908}) once re-referenced with sixteen compliance pairs, and \emph{collapses} to chance (\rcpt{0.564}) under its native \emph{safety} reference, the same failure the safety-trained guards show, further evidence that compliance and safety are distinct. ICS is the cheapest of these at the top of the range, reaching \rcpt{0.952} for one dot product on activations already computed (\rcpt{$\sim$0}\,ms) where GradSafe needs a forward\,$+$\,backward pass (\rcpt{200}\,ms) and a deployed guard a full autoregressive one, so on either cost axis it sits at the frontier, beaten only by GradSafe at 200ms and by nothing at its own (Figure~\ref{fig:pareto}, Table~\ref{tab:efficiency}).

At a fraction of the cost, and it adapts. ICS adds no trained parameters, and no extra forward pass \emph{when the scored text has already gone through the monitored model with its anchor-layer activation retained}; scoring externally supplied text costs one pass, like any first-party probe (Table~\ref{tab:efficiency}). It reaches full accuracy from \rcpt{20} pairs (\S\ref{sec:calib}) against the $10^4$--$10^5$ labelled examples a guard consumes. It is also \emph{recalibrable on the monitored model}: reading the monitored model's own activations, it re-fits in seconds on cached activations when that model is fine-tuned, where a shipped guard is a \emph{separate}, frozen network. A fresh ten-pair refit reads compliance at AUROC \rcpt{0.948}--\rcpt{0.962} on nine models spanning three architectures and multiple fine-tunes, all above the strongest deployed guard (LPG-4B \rcpt{0.943}); a verdict-tuned Llama-3.2-1B likewise keeps AUROC \rcpt{0.952}--\rcpt{0.957}, within \rcpt{0.007} of base across the tested fine-tunes (Table~\ref{tab:adapt}).

The paper's Comparison with Deployed Guards section reports two supporting checks in summary; the full tables are here. The first isolates whether ICS's edge over deployed guards is a ranking advantage or an artifact of guards running at an out-of-domain threshold. Every guard is given ICS's own target-domain threshold budget (5-fold within-domain, macro over 20 domains, no weight fine-tuning); the native-F1 collapse turns out to be a threshold artifact; a ten-pair re-threshold lifts every guard to F1 \rcpt{0.54}--\rcpt{0.80} (Table~\ref{tab:matchedcalib}). ICS's budget-free AUROC still clears every guard, so it leads every threshold metric even after the guards are given the same budget it gets.

\begin{table*}[htbp]
\caption{\textbf{Matched threshold calibration, the edge is ranking, not threshold access.} Every guard given ICS's target-domain threshold budget (5-fold within-domain, macro over 20 domains, no weight fine-tuning). The native-F1 collapse is a threshold artifact, a ten-pair re-threshold lifts guards to F1 \rcpt{0.54}--\rcpt{0.80}, but ICS's budget-free AUROC clears every guard, so it still leads every threshold metric at matched budget. AUROC for the four generative guards matches Table~\ref{tab:guardio}'s response mode, verified independently from the same raw per-case scores; ICS and the LLM judge have no input/response slot and are scored directly.}
\label{tab:matchedcalib}
\vspace{2mm}
\centering\small
\setlength{\tabcolsep}{10pt}
\fitwidth{%
\begin{tabular}{lrrrrrr}
\toprule
 & AUROC & \multicolumn{3}{c}{F1 at calibration budget} & FPR@95 & TPR@10 \\
\cmidrule(lr){3-5}
Arm & (rank, budget-free) & Native & 10-pair & Full-Data & (full-data) & (full-data) \\
\midrule
\textbf{ICS} (ours)   & \textbf{0.952} & \textbf{0.897} & \textbf{0.871} & \textbf{0.895} & \textbf{0.124} & \textbf{0.871} \\
Qwen3Guard-Gen (8B)   & 0.908 & 0.158 & 0.766 & 0.804 & 0.395 & 0.697 \\
LLM-judge (zero-shot, 8B) & 0.901 & 0.831 & 0.748 & 0.817 & 0.330 & 0.788 \\
WildGuard (7B)        & 0.875 & 0.124 & 0.691 & 0.715 & 0.593 & 0.527 \\
HarmBench cls (7B)    & 0.735 & 0.094 & 0.591 & 0.656 & 0.682 & 0.459 \\
Llama Guard 3 (1B)    & 0.749 & 0.265 & 0.542 & 0.580 & 0.766 & 0.330 \\
\bottomrule
\end{tabular}
}
\end{table*}

The second isolates the adaptation claim itself, that ICS re-fits to a fine-tuned model at zero cost while a shipped guard is a separate, frozen network. A fresh ten-pair refit is run on nine models spanning three architectures, five variants of Llama-3.1-8B (base plus medical, code, instruction, and safety fine-tunes) and the Gemma-3-12B and Qwen3-8B bases, each also with a finance fine-tune (Table~\ref{tab:adapt}). Every refit numerically exceeds the strongest deployed guard (LPG-4B, AUROC \rcpt{0.943}), which cannot adapt at all, though the margin is small on some arms (detailed below).

\begin{table*}[htbp]
\caption{\textbf{ICS re-fits across models and families, a guard cannot.} Each ICS row is a fresh ten-pair refit on that model's own activations, anchored and thresholded on disjoint validation; the guard column is identical for every protected model since it classifies the input case, not the activations. Across four Llama-3.1-8B fine-tunes ICS holds within \rcpt{0.007} of base; both cross-family bases hold within \rcpt{0.005} on their finance fine-tunes. Frozen-guard AUROCs credit each guard its stronger, response mode (Table~\ref{tab:guardio}). Every refit numerically exceeds the strongest deployed guard (LPG-4B \rcpt{0.943}), though the margin is as small as \rcpt{0.005} without a paired CI on that gap, read as parity-or-better, not a large win.}
\label{tab:adapt}
\vspace{2mm}
\centering\small
\setlength{\tabcolsep}{10pt}
\fitwidth{%
\begin{tabular}{llrr}
\toprule
Protected model & Family & \textbf{ICS AUROC$\uparrow$} & $\Delta$ vs.\ base \\
\midrule
\multicolumn{4}{l}{\textit{ICS refit on each model's own activations (0 params, 10 pairs)}} \\
Llama-3.1-8B-Instruct    & Llama & \textbf{0.962} & --- \\
\quad$+$ medical-SFT     & Llama & 0.958 & $-0.004$ \\
\quad$+$ code-SFT        & Llama & 0.962 & $0.000$ \\
\quad$+$ dolly-SFT       & Llama & 0.956 & $-0.006$ \\
\quad$+$ safe-align      & Llama & 0.962 & $0.000$ \\
Gemma-3-12B-it           & Gemma & 0.953 & --- \\
\quad$+$ finance-SFT     & Gemma & 0.948 & $-0.005$ \\
Qwen3-8B                 & Qwen  & 0.953 & --- \\
\quad$+$ finance-SFT     & Qwen  & 0.951 & $-0.002$ \\
\midrule
\multicolumn{4}{l}{\textit{Frozen guards (model-independent, cannot refit at all)}} \\
LPG-4B                   & ---   & 0.943 & fixed \\
LLM judge (zero-shot, 8B) & ---  & 0.912 & fixed \\
Qwen3Guard-Gen (8B)      & ---   & 0.908 & fixed \\
WildGuard (7B)           & ---   & 0.875 & fixed \\
HarmBench cls (7B)       & ---   & 0.735 & fixed \\
GLiGuard (300M)          & ---   & 0.703 & fixed \\
SIREN (14M probe)        & ---   & 0.697 & fixed \\
Llama Guard 3 (1B)       & ---   & 0.749 & fixed \\
\bottomrule
\end{tabular}
}
\end{table*}

\FloatBarrier
\section{Rule Ablation, Visualized}
\label{app:ruleablviz}
\begin{figure}[htbp]
\centering
\includegraphics[width=0.6\columnwidth]{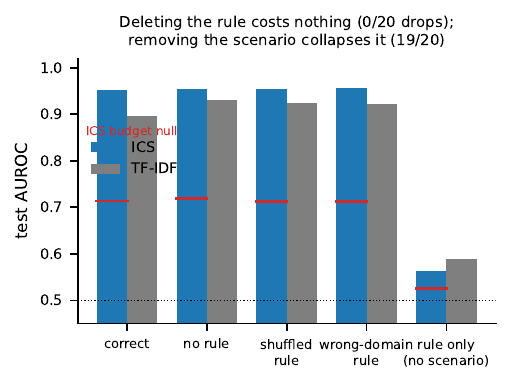}
\caption{\textbf{The rule is optional; the scenario carries the label.} ICS (blue) is unchanged whether the rule is correct, deleted, shuffled within-domain, or replaced by a different domain's rule (0/20 significant drops each); it collapses only when the \emph{scenario} is removed (rule-only, below its own null in 19/20). TF-IDF (grey) behaves the same, \emph{improving} when the rule is removed. Red bars are the ICS budget-matched null.}
\label{fig:ruleabl}
\end{figure}

The rule-ablation result of the paper's What the Score Measures section, drawn rather than tabulated. Figure~\ref{fig:ruleabl} puts ICS, the TF-IDF floor, and the budget-matched null on one axis across all five rule conditions, so the flat profile under rule corruption and the single collapse under scenario removal can be read off at a glance.

\subsection{Benign Inputs (Over-Flagging)}
\label{app:benign}

\begin{table*}[htbp]
\caption{\textbf{Benign inputs (over-flagging).} Fraction of clearly-benign inputs flagged as violation, lower is better ($N{=}250$/1319/1000). ICS is worse than well-calibrated guards on easy benign (XSTest 0.44, Alpaca 0.18) but lower on adversarially-benign OR-Bench (\rcpt{0.327}) than every guard that fires on compliance at all. Three arms score numerically lower still (Llama Guard 3 \rcpt{0.000}, HarmBench \rcpt{0.005}, Llama Guard 3~8B \rcpt{0.193}), but as refusal classifiers that essentially never fire on OR-Bench's style regardless of content, a floor artifact, not better calibration; read this beside detection performance (Table~\ref{tab:metrics}).}
\label{tab:benign}
\vspace{2mm}
\centering\small
\setlength{\tabcolsep}{16pt}
\fitwidth{%
\begin{tabular}{lrrr}
\toprule
Arm & XSTest-safe FPR & OR-Bench-hard FPR & Alpaca/Dolly FPR \\
\midrule
\textbf{ICS} (ours) & 0.437 & 0.327 & 0.182 \\
MLP probe (same activations) & 0.393 & 0.355 & 0.224 \\
Qwen3Guard-Gen (8B) & 0.056 & 0.568 & 0.001 \\
LLM judge (zero-shot, 8B) & 0.176 & 0.533 & 0.025 \\
WildGuard (7B) & 0.012 & 0.751 & 0.003 \\
SIREN & 0.840 & 0.836 & 0.409 \\
Llama Guard 3 (1B) & 0.000 & 0.000 & 0.000 \\
HarmBench cls (7B) & 0.028 & 0.005 & 0.000 \\
GLiGuard & 0.172 & 0.560 & 0.016 \\
Llama Guard 3 (8B) & 0.032 & 0.193 & 0.002 \\
TF-IDF floor & 0.406 & 0.406 & 0.425 \\
\bottomrule
\end{tabular}
}
\end{table*}

\clearpage
\FloatBarrier
\section{Selection Nulls: the Budget-Matched Detection Floor}
\label{sec:nulls}

\noindent\textbf{Layer selection alone lifts a random direction to 0.71.} Layer selection is not free. Shuffling labels within pairs leaves a random-signed $O(1/\sqrt{n})$ component of the true direction in every layer; taking the argmax over layers on a validation slice selects that leak. A random unit direction pushed through ICS's own pipeline (argmax over 17 layers) scores \rcpt{0.7135} on average, not 0.5, and a random direction given the refusal direction's larger budget scores \rcpt{0.7747}; three independent experiments agree on this floor to within 0.004. ICS is accordingly reported against this budget-matched null throughout, not against chance: AUROC \rcpt{0.952} against a null of \rcpt{0.714}. We do not visualise or tabulate this margin against other arms' margins, since each arm faces a different search space and a different null construction, and a longer margin does not identify a better method; the full per-arm breakdown is released with the evaluation protocol.

Three consequences. First, \emph{absolute} probe numbers are inflated: ``ICS achieves 0.95'' is a statement made against a floor of 0.71, not 0.50. Second, the refusal direction, fit with zero compliance supervision, retains a genuine but small margin ($+0.089$; tail-significant against matched random directions in only 6--8 of 20 domains). Its 0.863 is therefore \emph{not} evidence that generic harmfulness carries most of the compliance signal: against budget-matched nulls, harmfulness accounts for roughly \rcpt{37\%} of ICS's honest margin, and per-domain calibration for the majority. Third, the zero-selection arms show the largest margins over their own nulls. We resist the obvious reading. Those arms search nothing, so their null is chance, while a layer-selected direction is scored against \rcpt{0.714}; the two margins are differences from different reference points and a larger one does not identify a better method. What the comparison does establish is narrower and still uncomfortable: the activation probe's advantage over a bag of words is not visible once each is read against its own budget, so the case for ICS rests on cost and first-party access rather than on a margin the null structure cannot support.

The surface-residualized score survives the same correction: after projecting out lexical content, sentiment, length, and entropy, ICS scores \rcpt{0.771} against a measured budget-matched null of \rcpt{0.587} (residualization removes part, not all, of the selection leak), above it in 18/20 domains. The activations carry \emph{some} signal beyond the text surface; the practical margin over a surface model is what fails (\S\ref{sec:ablations}).

\FloatBarrier
\section{Selection-Null Transfer Beyond Compliance}
\label{app:generality}
The budget-matched null construction used throughout this paper (\S\ref{sec:nulls}) is not a compliance-specific artifact: the same layer-selection procedure applied to unrelated published probing settings (truthfulness, sentiment) inflates a random direction to \rcpt{0.50--0.65} against a no-selection control of \rcpt{0.49--0.51}, confirming the null is measuring a real property of layer selection rather than something particular to this task. Compliance's \rcpt{0.714} is the highest floor measured of these settings. Full per-setting breakdown, the $n$-decay check, and null-generation detail are released with the evaluation protocol.

\FloatBarrier
\section{Content-Selection: Generated Responses and the Judged Regulatory Task}
\label{app:contentsel}
The main text (the paper's Score-Guided Candidate Selection section) establishes verdict-selection on a classification corpus. Two extensions test whether the gain survives when the candidates are substantively different generated responses rather than menu labels (Table~\ref{tab:contentsel-main}).

\begin{table}[t]
\caption{\textbf{Content-selection survives on generated text, both mechanically-verified and judged.} \emph{IFEval}, ICS-guided pick of $k$ responses vs.\ the instruction-following verifier (CI $[+2.4,+7.8]$). \emph{Regulatory}, frozen ICS direction on LLM-judged advisory responses (CI $[+8.8,+14.2]$, all six domains significant).}
\label{tab:contentsel-main}
\vspace{2mm}
\centering\small
\setlength{\tabcolsep}{6pt}
\fitwidth{%
\begin{tabular}{lrrrr}
\toprule
Task & $n$ & Vs.\ Random & Vs.\ First & Oracle Headroom \\
\midrule
IFEval (mechanical) & 541 & \textbf{+5.2}pp & +6.1pp & 22.6\% \\
Regulatory (LLM judge) & 6 dom., $k{=}5$ & \textbf{+11.5}pp & --- & 43\% \\
\bottomrule
\end{tabular}
}
\end{table}

\paragraph{It also works on generated content, not just verdicts.} The deployment case (generate $k$ substantively different responses, return the most compliant) we test on IFEval, whose mechanical \texttt{instruction\_following} verifier gives objective per-response ground truth (no judge). Over all \rcpt{541} prompts, ICS-guided content-selection raises the strictly-verified pass rate by \rcpt{$+5.2$}pp over random (CI $[+2.4,+7.8]$, $p{=}1/2000$) and \rcpt{$+6.1$}pp over first-pick, replicating on a held-out block (\rcpt{$+4.5$}pp, $p{=}0.036$); on the \rcpt{237} prompts where candidates actually vary, the gain is \rcpt{$+11.8$}pp. The candidates are diverse (mean pairwise word-dissimilarity \rcpt{0.91}), so this is content-selection, not relabelling. The effect is modest, \rcpt{22.6\%} of oracle headroom, ceilinged by a within-prompt detection AUROC of \rcpt{0.62}, but positive, powered, and mechanically verified. It is not a 1B artifact: content-selection raises the IFEval verified pass rate on \rcpt{4} of 5 models (\rcpt{$+2.4$} to \rcpt{$+5.2$}pp), each scaling with within-prompt detection AUROC (\rcpt{0.60--0.63}); it is null only on Qwen2.5-7B, whose within-prompt detection collapses to chance (\rcpt{0.53}). The gain is gated by whether ICS separates a prompt's compliant candidates from its violating ones, not by headroom, since content detection is the harder problem.

\paragraph{And on the regulatory domain itself, judged.} IFEval measures format compliance; the paper's subject is regulatory compliance, where no mechanical verifier exists. We therefore generate advisory responses to violating scenarios (six domains, $k{=}5$) and score them with an LLM judge. The selector is the \emph{frozen} receipted ICS direction, never fit on judge labels or generated text, so there is no selector--judge circularity. Guided selection raises the judged compliance rate by \rcpt{$+11.5$}pp pooled (CI $[+8.8,+14.2]$, $p{=}0.001$, all six domains significant), capturing \rcpt{43\%} of oracle headroom (\rcpt{$+16.7$}pp on varying prompts). A coherence control rules out quality-selection: judged compliance is uncorrelated with length (\rcpt{$-0.06$}) and ICS is if anything \emph{negatively} length-correlated, and restricting to coherent candidates leaves the gain unchanged. The judge is a fallible oracle, so this bounds rather than proves; but across a mechanical verifier and a judge, on format and on regulation, content-selection is a real, modest, deployable gain. This claim, and the separate rule-counterfactual claim of the paper's What the Score Measures section, are both cross-checked against three independent frontier judges in \S\ref{app:frontierjudge}.

\FloatBarrier
\section{Independent Frontier-Model Validation of the LLM-Judge Claims}
\label{app:frontierjudge}
\begin{figure}[t]\centering
\includegraphics[width=\columnwidth]{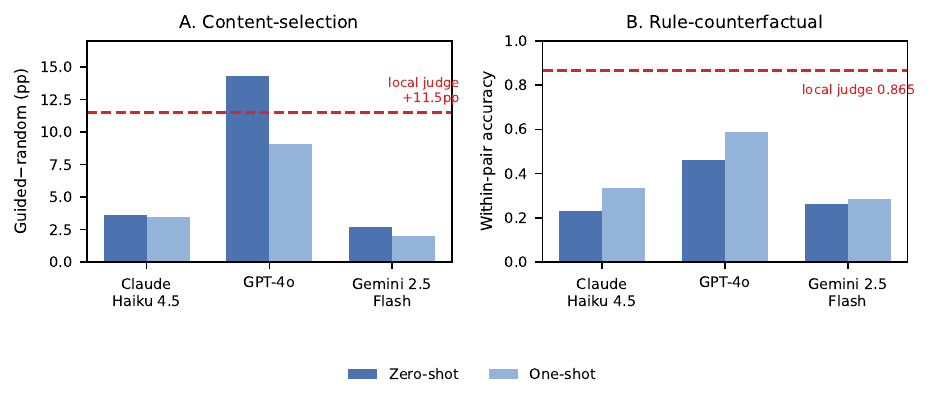}
\caption{\textbf{Frontier judges mostly confirm the local judge's weaknesses, not just its strengths.} \emph{Left}, content-selection guided-random gap (pp) vs.\ the local judge's \rcpt{+11.5}pp (dashed). \emph{Right}, rule-counterfactual within-pair accuracy vs.\ the local judge's \rcpt{0.865} (dashed).}
\label{fig:judgevalidation}
\end{figure}

Both LLM-judge claims above rest on a single zero-shot judge, Llama-3.1-8B-Instruct, so we cross-check both against three independent judges (Claude Haiku 4.5, GPT-4o, Gemini 2.5 Flash) under exact-prompt replication of the local judge's own protocol, zero-shot and one-shot (Table~\ref{tab:judgesens}, Figure~\ref{fig:judgevalidation}). All three judges preserve the \emph{direction} of the content-selection effect (guided $>$ random $>$ first) but its magnitude varies by an order of magnitude across judges. Rule-counterfactual agreement is weaker and should be treated as judge-sensitive: only one configuration (GPT-4o, one-shot) closes most of the gap to the local judge; the other five stay well below it. This rule-flip metric is measured on the fixed-scenario counterfactual set (\S\ref{app:fixedscenariocf}), not the confound-free crossed benchmark, and that set has its own known confound, rule polarity alone predicts the label (\S\ref{app:crossed}); a judge that under-performs the local model here is not shown to be more rule-blind, only that it is more or less exploitable by the same polarity cue. The reading we do not draw is that three low frontier scores make the local judge's \rcpt{0.865} the confirmed number: with only four judges and a confounded task, this cross-check establishes that the direction of the content-selection effect is judge-general, and flags rule-counterfactual magnitude as judge-sensitive and requiring the unconfounded crossed benchmark (\S\ref{app:crossed}), not the local judge alone, as the basis for the paper's rule-blindness claim. Full per-domain, per-shot breakdown and the earlier flagship-judge round are released with the evaluation protocol.

\begin{table}[t]
\caption{\textbf{Judge sensitivity, summarised.} Guided$-$random is the content-selection gap; rule-flip accuracy is within-pair accuracy on the 416-row rule-flip set. Full six-configuration table released with the evaluation protocol.}
\label{tab:judgesens}
\vspace{2mm}
\centering\small
\setlength{\tabcolsep}{6pt}
\fitwidth{%
\begin{tabular}{lrrr}
\toprule
Judge (shot) & Guided$-$random & Rule-flip acc. & Unparseable \\
\midrule
Local judge & \textbf{+11.5}pp & 0.865 & --- \\
GPT-4o (zero) & +14.3pp & 0.462 & low \\
GPT-4o (one) & +9.1pp & \textbf{0.586} & 69/416 \\
Claude Haiku (zero) & +3.6pp & 0.229 & low \\
Gemini 2.5 (zero) & +2.7pp & 0.263 & low \\
\bottomrule
\end{tabular}
}
\end{table}

\FloatBarrier
\section{Adversarial Robustness: Verdict- vs.\ Content-Selection}
\label{app:attack}
The compliant-register suffix that flips the monitor in \S\ref{sec:ablations} is applied to the selection pipeline (1B, 20 domains, $k{=}5$; the selector scores the case, so a suffix that flips the case verdict also flips the selector). Ground truth is the true verdict, since the cosmetic suffix does not change actual compliance. Table~\ref{tab:attack} reports captured headroom (guided${-}$random)/(oracle${-}$random) under no attack, a uniform suffix on all $k$ candidates, and a poison suffix on the truly-violating candidates only (the untrusted-generator / content-filtering threat model). The mechanism is the monitor attack, downstream: the suffix flips the case verdict VIOLATE$\to$COMPLY on \rcpt{70.8\%} of violating cases (\rcpt{81.5\%} among clean-flagged violations), so the selector's selected-COMPLY rate on violating cases rises \rcpt{$0.10\to0.74$} and true pass on those cases falls \rcpt{$0.90\to0.26$}. Selection survives only in the \rcpt{3}/20 domains where the monitor also resists; \rcpt{11}/20 go negative. Because selection inherits the monitor's case verdict, it has no floor above random once the monitor is defeated.

\begin{table*}[htbp]
\caption{\textbf{The attack defeats \emph{verdict}-selection.} Captured headroom collapses from 78.8\% to the random floor under a poison suffix; guided true-pass drops to 0.552, essentially random (0.541). Verdict-selection inherits the monitor's case verdict and fails wherever the monitor fails, but content-selection does not (below).}
\label{tab:attack}
\vspace{2mm}
\centering\small
\setlength{\tabcolsep}{16pt}
\fitwidth{%
\begin{tabular}{lrrr}
\toprule
Condition & Headroom captured & Guided true-pass & $\Delta$ vs.\ baseline \\
\midrule
(A) baseline (no attack) & \textbf{0.788} & 0.872 & --- \\
(B) uniform suffix (all $k$) & 0.095 & 0.581 & $-0.291$ \\
(C) poison (violating cases only) & 0.027 & 0.552 & $-0.320$ \\
\midrule
\multicolumn{4}{l}{\textit{random-selection floor $=0.541$; poison headroom CI $[-0.014,+0.067]$ (includes 0);}} \\
\multicolumn{4}{l}{\textit{$\Delta$ true-pass CI $[-0.335,-0.305]$, $p=1/10000$ (both B and C).}} \\
\bottomrule
\end{tabular}
}
\end{table*}

\paragraph{Content-selection resists the same attack.} Applied to the deployable content-selection lead (IFEval, 1B, $k{=}5$, official mechanical verifier; the poison design excludes the \rcpt{9.4\%} of candidates whose strict verdict the suffix genuinely flips, so we measure only ICS being fooled by register into selecting a still-failing response), the attack moves the verified pass rate by \rcpt{$-1.1$}pp (CI $[-0.033,+0.011]$, incl.\ 0, $p{\approx}0.36$) and retains \rcpt{79\%} of headroom (\rcpt{$0.226\to0.177$}); a uniform register shift does not reorder the ranking. The mechanism is measured: the suffix moves a response's ICS by only \rcpt{$+0.10$} versus \rcpt{$+2.2$} for the case-text attack (${\sim}22\times$ smaller), because the scored last token follows a diverse ${\sim}1000$-token response that a $13$-token boilerplate tail cannot dominate, and right-truncation removes the suffix on \rcpt{19\%} of candidates. Even a selective adversary's harm is bounded at ${\sim}\rcpt{6.4}$pp. The deployable use is robust for the same reason the monitor is fragile, last-token locality, but read the other way: scoring long, diverse content dilutes the local perturbation that dominates a short case text. \paragraph{But an adaptive white-box attack breaks it.} The fixed suffix is non-adaptive. A GCG \citep{zou2023adversarial} suffix optimised \emph{directly} against the ICS direction (white-box; 12 tokens, 40 steps; 20 stratified attackable IFEval prompts, 1B) is far stronger: it moves a response's faithful ICS by \rcpt{$+1.66$} versus \rcpt{$+0.10$} for the fixed suffix (${\sim}16\times$) while the official verifier \emph{still fails} 52/53 poisoned responses, driving ICS-guided verified pass from \rcpt{0.70} to \rcpt{0.00} (captured headroom inverts below random; $\Delta$ \rcpt{$-0.70$}, CI $[-0.90,-0.50]$, $p{=}1/2000$). Content selection's robustness is therefore bounded to the \emph{non-adaptive} threat model: against a white-box adversary optimising against the score itself, the ICS content-monitor is not safe as a sole selector. (Subset of 20 stratified attackable prompts; the collapse to \rcpt{0.00} verified pass on every poisoned response is a floor, not a marginal effect, so the direction of the finding does not depend on sample size.)


\FloatBarrier
\section{Ablations and Robustness}
\label{sec:ablations}

\subsection{Calibration Sensitivity: which ten pairs, and how many?}
\label{sec:calib}
\begin{figure}[htbp]\centering
\begin{minipage}{0.5\textwidth}\centering
\includegraphics[width=0.96\linewidth]{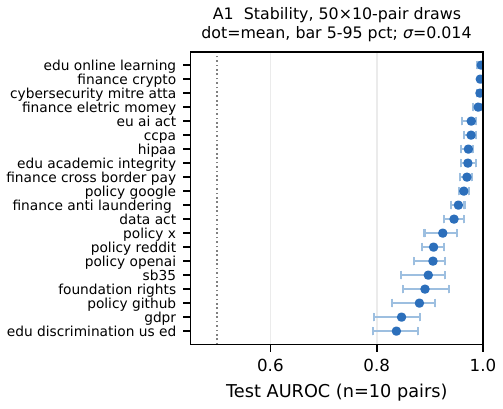}
\end{minipage}%
\begin{minipage}{0.5\textwidth}\centering
\includegraphics[width=0.96\linewidth]{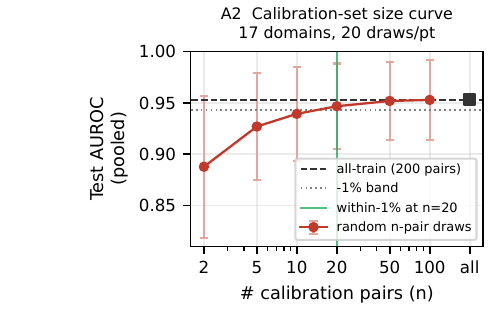}
\end{minipage}\\[2mm]
\begin{minipage}{0.5\textwidth}\centering
\includegraphics[width=0.96\linewidth]{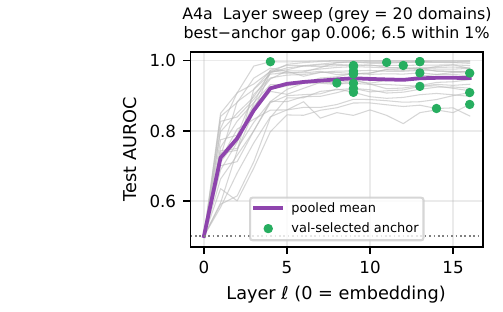}
\end{minipage}%
\begin{minipage}{0.5\textwidth}\centering
\includegraphics[width=0.96\linewidth]{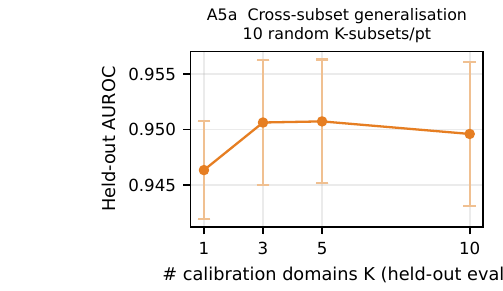}
\end{minipage}
\caption{\textbf{ICS is insensitive to the calibration set.} \emph{A1 (top left):} test AUROC across 50 random 10-pair calibration draws per domain (dot = mean, bar = 5th--95th percentile; tight across domains, $\sigma\approx0.014$). \emph{A2 (top right):} AUROC vs.\ number of calibration pairs $n$, within 1\% of the full-data value by $n{=}20$; calibration on \textsc{train} only, evaluated on the disjoint test split. \emph{A4a (bottom left):} test AUROC by candidate anchor layer for all 20 domains (grey), the pooled mean (purple), and each domain's validation-selected anchor (green); an intermediate anchor-stability variant was consolidated into this panel and is omitted. \emph{A5a (bottom right):} held-out AUROC vs.\ the number of calibration domains $K$ pooled before fitting the direction, 10 random $K$-subsets per point.}
\label{fig:calib}
\end{figure}

A training-free guard invites the question: you say ten pairs, but \emph{which} ten, and must they be in-domain? We vary only the calibration set (pairs feeding the direction), holding the layer anchor on \textsc{val-layer} and the threshold on \textsc{val-thresh}, and report on the disjoint test split ($\text{train}\cap\text{test}=\emptyset$ asserted per draw). Three answers (Figure~\ref{fig:calib}). \textbf{(1) Which ten is stable:} over 50 random 10-pair draws per domain the mean within-domain AUROC standard deviation is \rcpt{0.014} (F1 \rcpt{0.019}); a practitioner drawing ten pairs at random lands within about $\pm\rcpt{0.08}$ AUROC of the typical result. \textbf{(2) Ten pairs is near enough:} AUROC rises $\rcpt{0.888}$ (n{=}2) $\to \rcpt{0.939}$ (n{=}10) $\to \rcpt{0.947}$ (n{=}20) and plateaus, reaching within \rcpt{1\%} of the full-data value by \rcpt{20} pairs, against the $10^4$--$10^5$ labelled examples a trained guard consumes. \textbf{(3) In-domain barely matters:} a single \emph{generic} direction fit from the pooled training pairs scores \rcpt{0.953}, versus \rcpt{0.952} for per-domain calibration and \rcpt{0.943} for blind cross-domain transfer, so one direction can ship for all domains at a $-\rcpt{0.001}$ cost. \textbf{(4) The calibrate-once LODO headline needs roughly ten pairs \emph{per source distribution}, not ten total.} The 0.728 mean leave-one-distribution-out AUROC (the paper's Generalisation Beyond the Calibration Distribution section) is fit on the code's actual pooled train budget, not a literal ten-pair draw; we re-fit the identical LODO protocol under four explicit training-budget conditions, three random draws each, same held-out val for anchor and threshold throughout. Ten pairs drawn from the \emph{entire} pool, total, collapses toward chance (mean AUROC \rcpt{0.556}$\pm$\rcpt{0.050}); twenty pairs total fares little better (\rcpt{0.615}$\pm$\rcpt{0.041}); ten pairs drawn from \emph{each} source distribution (roughly 100 pooled total across the ten held-in distributions) recovers most of the headline (\rcpt{0.707}$\pm$\rcpt{0.023}); the full pool (the code's actual, larger budget) reaches the reported \rcpt{0.728}. The near-zero-cost framing is accurate for \emph{per-distribution} sample size, not for the total labelled-pair count, and we state it that way in the main text accordingly.

\paragraph{Template robustness.} The single-template concern is answered along the axis that matters. Presenting the same rule and scenario under a paraphrased instruction, a terse one-liner, or the model's native chat template leaves ICS within \rcpt{0.947}$\pm\rcpt{0.008}$ AUROC and preserves \rcpt{96\%} of cross-template transfer, so the direction is not an artifact of OmniCompliance's exact wording. It is sensitive along one axis, \emph{position}: reordering so the scored text ends on the (generic, truncated) rule rather than the case narrative collapses detection to \rcpt{0.668} and breaks transfer, because ICS is a last-token probe of the scenario's generic violation signal (the paper's What the Score Measures section). The operating condition is that the scored text end on the case, not that the wording be fixed.

\noindent\textbf{The direction is largely one axis, and this is not a 1B artifact.} The 20 per-domain directions are substantially \emph{one} direction, consistent with the calibration finding that a single generic direction generalises (\S\ref{sec:calib}) and with reading a shared violation-register axis rather than per-rule geometry. Table~\ref{tab:geomcross} replicates every quantity on two more models. \textbf{The collapse itself is real on all three}, but its \emph{degree} is not identical: the 8B-scale arms, Llama and Qwen alike, spread over a few more effective dimensions (participation ratio \rcpt{3.2}--\rcpt{3.5}) than the 1B arm (\rcpt{1.47}) -- consistent with scale, not family, and we report the pattern rather than rounding it to ``the same everywhere.'' The positive cosine to the refusal direction and the near-zero LODO-pooling cost both hold on every model.

\begin{table*}[htbp]
\caption{\textbf{Direction collapse replicates across scale and family.} All AUROC gaps are small (collapse); all cosines to the refusal direction are positive and far from the random-vector null; LODO pooling never costs more than \rcpt{0.01} AUROC on any model. The one real difference: both 8B arms are less extremely collapsed than the 1B arm (participation ratio ${\sim}3.2$--$3.5$ vs.\ \rcpt{1.47}), a scale-consistent pattern across two families, not a single-model quirk.}
\label{tab:geomcross}
\vspace{2mm}
\centering\small
\setlength{\tabcolsep}{10pt}
\fitwidth{%
\begin{tabular}{lrrrr}
\toprule
Model & Cross-domain AUROC gap & Participation ratio & Cosine to refusal direction & LODO $\Delta$ \\
 & (diag$-$off-diag) & (random-vector null \rcpt{18.0}) & (std.\ deviations) & (vs.\ per-domain) \\
\midrule
Llama-3.2-1B & \rcpt{+0.0085} & \rcpt{1.47} & \rcpt{+0.151} (\rcpt{6.8}$\sigma$) & \rcpt{+0.0005} \\
Llama-3.1-8B & \rcpt{+0.0120} & \rcpt{3.52} & \rcpt{+0.119} (\rcpt{7.6}$\sigma$) & \rcpt{+0.0091} \\
Qwen3-8B & \rcpt{+0.0195} & \rcpt{3.24} & \rcpt{+0.09--0.13} (\rcpt{5.8--8.2}$\sigma$) & \rcpt{+0.0066} \\
\bottomrule
\end{tabular}
}
\end{table*}

An OLS residualization against lexical content, sentiment, length and entropy leaves a significant \rcpt{0.771} (null \rcpt{0.587}, above it in 18/20), so activations do carry signal beyond the surface. We do not claim it exceeds a strong lexical model on this single-template benchmark, which is why the deployable claims rest on selection and the guard comparison, not on an absolute detection margin.

\noindent\textbf{Design choices are swept and secondary} (\S\ref{app:axes}, Table~\ref{tab:axes}). Raw and cosine scoring tie ($\Delta \le 0.001$) and $\lVert a\rVert$ alone is near chance, so no length confound rides the norm; last-token/residual/mean is the weakly-best cell of the pooling/site/estimator grid; the equal-variance threshold assumption is false, so we report AUROC as primary. Three perturbations locate what the probe reads: a \emph{prepended} opposite-verdict sentence moves nothing (\rcpt{0.012}), paraphrase is its best cell (retention \rcpt{0.994}, beating TF-IDF in 18/20), but a single \emph{appended} compliant-register sentence flips \rcpt{72.4\%} of caught violations (TF-IDF \rcpt{3.4\%}): the probe reads recent register at the read position, not content (\S\ref{app:perturb}).

\FloatBarrier
\section{Metric Robustness}
\label{app:metric}
Every headline detection finding is re-expressed on the same seed-42 splits (positive $=$ violation) under AUROC, AUPRC, $\mathrm{FPR}@95\%\mathrm{TPR}$, and $\mathrm{TPR}@10\%\mathrm{FPR}$; all 100 re-extracted rule conditions match the stored rule-ablation AUROC to $<10^{-9}$ before scoring. Table~\ref{tab:metric} shows the qualitative conclusions do not depend on the metric. Rule-blindness is the decisive case: deleting, deranging, or swapping the rule causes \textbf{0/20} BH-significant degradations under \emph{all four} metrics ($\mathrm{FPR}@95$ moves $0.176\to0.182$), while the rule-only positive control collapses \rcpt{19/20} under all four. AUPRC is a weak stress test here by construction (prevalence $0.5$ makes its baseline $0.5$, so it tracks AUROC and on 5/20 domains \emph{understates} ICS); we therefore lead the deployment claims with $\mathrm{FPR}@95\%\mathrm{TPR}$, where ICS's advantage over the same lexical floor, now under this metric, is both largest (\rcpt{0.164} vs.\ \rcpt{0.426}) and operationally meaningful.

\begin{table*}[htbp]
\caption{\textbf{The findings are metric-robust.} ICS's floor-beating margin, the rule-blindness null, and the rule-only positive control all hold under four metrics, so the central results are not AUROC artifacts. Calibration: within-domain Platt ECE 0.023, cross-domain 0.156 (\S\ref{app:thresh}).}
\label{tab:metric}
\vspace{2mm}
\centering\small
\setlength{\tabcolsep}{18pt}
\fitwidth{%
\begin{tabular}{lrrrr}
\toprule
Finding (domains of 20 supporting it) & AUROC & AUPRC & FPR@95\%TPR & TPR@10\%FPR \\
\midrule
\textbf{ICS beats lexical floor} & 18 & 15 & 18 & 17 \\
Rule-deletion BH-sig.\ \emph{drops} (should be 0) & \textbf{0} & \textbf{0} & \textbf{0} & \textbf{0} \\
Rule-only positive control collapses & 19 & 19 & 19 & 19 \\
\bottomrule
\end{tabular}
}
\end{table*}
\clearpage
\FloatBarrier
\section{Threshold Transfer: Direction vs.\ Operating Point}
\label{app:thresh}
The same failure shows up as miscalibration: a Platt-scaled ICS is well-calibrated within domain (ECE \rcpt{0.023}) but not across (cross-domain ECE \rcpt{0.156}, p95 \rcpt{0.40}), so a fixed cross-domain $\tau$ inherits exactly this (\S\ref{app:metric}). We fit the ICS direction and threshold $\tau$ per domain and evaluate every ordered domain pair (1B, 20 domains, scoring-only over index-identical splits; each domain's stored receipt AUROC reproduced to $<0.002$ before transfer). The readout \emph{direction} is portable; the \emph{operating point} is not (Table~\ref{tab:thresh}). A $\tau$ calibrated to 5\% FPR on one domain is heavy-tailed elsewhere (median 7\%, but 26--31\% at p95), and $\mathrm{FPR}@95\%\mathrm{TPR}$ blows up from 27.7\% within-domain to a cross-domain mean of 34.7\% and 77--94\% at p95, with recall down to 25\%. Midpoint-$\tau$ accuracy hides this (mean $0.903\to0.880$) but collapses to $\approx$chance (0.556) in the tail. The direction that identifies compliance transfers; the threshold that decides it must be re-calibrated per domain.

\begin{table*}[htbp]
\caption{\textbf{The direction transfers; the operating point does not.} Cross-domain AUROC nearly matches within-domain (all 380 ordered pairs $\ge0.79$), but a fixed threshold's realized FPR is heavy-tailed and $\mathrm{FPR}@95\%\mathrm{TPR}$ blows up 3--4$\times$. Distribution reported via p95 and a small-$n$-robust 210-pair subset, not the worst cell.}
\label{tab:thresh}
\vspace{2mm}
\centering\small
\setlength{\tabcolsep}{20pt}
\fitwidth{%
\begin{tabular}{lrr}
\toprule
Quantity & Within-domain & Cross-domain \\
\midrule
Detection AUROC (the direction) & 0.952 & 0.943 \\
\quad(all 380 ordered domain pairs) & --- & $\ge 0.79$ \\
$\mathrm{FPR}@95\%\mathrm{TPR}$ & 0.277 & 0.347 (p95 0.77--0.94) \\
Realized FPR of a 5\%-calibrated $\tau$ & 0.050 & median 0.07 (p95 0.26--0.31) \\
Accuracy at the midpoint $\tau$ & 0.903 & 0.880 (worst-case tail 0.556) \\
\bottomrule
\end{tabular}
}
\end{table*}

\FloatBarrier
\section{Perturbation Robustness}
\label{app:perturb}
Three perturbations of the frozen probe. A \emph{prepended} opposite-verdict sentence moves almost nothing (swap sensitivity \rcpt{0.012}, the lowest of any arm). Under \emph{paraphrase}, ICS retains \rcpt{0.994} with zero significant losses, degrading less than TF-IDF in \rcpt{18/20} domains. But one \emph{appended} compliant-register sentence flips \rcpt{72.4\%} of correctly-flagged violations, while TF-IDF moves \rcpt{3.4\%}.

\FloatBarrier
\section{Cross-Model Replication}
\label{sec:models}

ICS's margin over TF-IDF replicates across all \textbf{twelve} arms we ran, spanning \textbf{four} families and \textbf{1B to 72B}, inside a narrow band from \rcpt{$+0.053$} (Qwen3.5-4B) to \rcpt{$+0.068$} (Qwen2.5-7B and Qwen2.5-72B, tied), without a consistent scaling trend: margin rises with scale \emph{within} the Llama family but the trend does not survive crossing families (Table~\ref{tab:crossmodel}). \textbf{The 70B-class question is no longer open}: Qwen2.5-72B-Instruct (int8, single GPU) lands at margin \rcpt{$+0.068$}, inside the established band, not above it. Layer selection is performed separately per model, since the signal sits at a consistent \emph{relative} rather than absolute depth; raw and cosine scoring remain similar across every model ($\Delta \le 0.001$). One family (Qwen3.5) is a named exception to the relative-depth pattern with no mechanistic account offered; full anchor-position trends, per-domain profile correlations, and norm-ablation detail are released with the evaluation protocol.

\begin{table*}[t]
\caption{\textbf{The finding replicates across four architecture families and two orders of magnitude in scale (1B--72B).} Margin $=$ mean ICS$_{\text{raw}}$ AUROC $-$ mean TF-IDF-floor AUROC (\rcpt{0.896}), outcome-ablated OmniCompliance, 20 domains, identical procedure for every row. An earlier-draft per-model ``gate'' column has been withdrawn as not the registered gate of \S\ref{sec:gate}; provenance released with the evaluation protocol. \texttt{google/gemma-3-4b-it} excluded: gated repository at submission time. Llama-3.3-70B and Qwen2.5-72B are the two frontier-scale arms (int8); Qwen3.5-27B replicates the margin band but is a named exception to the cross-family anchor-depth band (\S\ref{sec:models}).}
\label{tab:crossmodel}
\vspace{2mm}
\centering\small
\setlength{\tabcolsep}{9pt}
\fitwidth{%
\begin{tabular}{lrrrrrr}
\toprule
Model & Family & Params & $n_{\text{layers}}$ & Mean ICS$_{\text{raw}}$ & Mean ICS$_{\text{cos}}$ & Margin vs.\ TF-IDF floor \\
\midrule
Llama-3.2-1B-Instruct (reference) & Llama & 1B & 16 & 0.952 & 0.953 & \textbf{+0.056} \\
Llama-3.2-3B-Instruct & Llama & 3B & 28 & 0.958 & 0.959 & \textbf{+0.062} \\
Llama-3.1-8B-Instruct & Llama & 8B & 32 & 0.962 & 0.963 & \textbf{+0.067} \\
\textbf{Llama-3.3-70B-Instruct} & Llama & \textbf{70B} & \textbf{80} & \textbf{0.955} & \textbf{0.956} & \textbf{+0.059} \\
Mistral-7B-Instruct-v0.3 & Mistral & 7B & 32 & 0.961 & 0.961 & \textbf{+0.066} \\
Gemma-2-9B-it & Gemma & 9B & 42 & 0.959 & 0.960 & \textbf{+0.063} \\
Qwen2.5-7B-Instruct & Qwen & 7B & 28 & 0.963 & 0.964 & \textbf{+0.068} \\
Qwen3-8B & Qwen & 8B & 36 & 0.953 & 0.957 & \textbf{+0.057} \\
Qwen3.5-4B & Qwen & 4B & 32 & 0.948 & 0.949 & \textbf{+0.053} \\
Qwen3.5-9B & Qwen & 9B & 32 & 0.951 & 0.951 & \textbf{+0.055} \\
Qwen3.5-27B & Qwen & 27B & 64 & 0.952 & 0.953 & \textbf{+0.056} \\
\textbf{Qwen2.5-72B-Instruct (int8)} & Qwen & \textbf{72B} & \textbf{80} & \textbf{0.964} & \textbf{0.965} & \textbf{+0.068} \\
\bottomrule
\end{tabular}
}
\end{table*}

\paragraph{The 72B Arm: Methodology}
\label{app:scale72b}
Llama-3.3-70B-Instruct and Qwen2.5-72B-Instruct were evaluated in int8 (\texttt{BitsAndBytesConfig(load\_in\_8bit=True)}) because the released bf16 checkpoints did not fit a single 97\,GB GPU ($\approx$144\,GB of weights); both fit with headroom and needed no multi-GPU sharding, and the same seed-42 split, direction fit, and per-domain layer search were used as every other arm in Table~\ref{tab:crossmodel}, only the quantised forward pass differs.

\FloatBarrier
\section{Method Axes}
\label{app:axes}
Pooling $\times$ probe site $\times$ centroid estimator, swept around the published configuration (paired bootstrap against it, shared resample indices, exact-arithmetic BH). Pooling matters and last-token is right, and not because of truncation, since no OmniCompliance text exceeds 512 tokens: mean-over-all-tokens costs \rcpt{$-0.099$} (15/20 significant losses) and falls into the selection-null zone in \rcpt{9/20} domains; max-pooling costs \rcpt{$-0.067$}; response-only mean \rcpt{$-0.020$}. Probe site is a tie: attention output trends \rcpt{$+0.004$} (positive in 15/20, never BH-significant), MLP output is a wash. Robust centroid estimators (median, 20\% trimmed) change nothing ($|\Delta| \le \rcpt{0.0007}$).

\begin{table*}[htbp]
\caption{Method-axis sweep, mean over 20 ablated domains (paired bootstrap vs.\ the published cell, exact-BH). Last-token/residual/mean is on the Pareto face; attention output is a non-significant $+0.004$; the centroid estimator is a no-op. Mean-over-all-tokens is the only variant that falls into the selection-null zone (11/20 clear it).}
\label{tab:axes}
\vspace{2mm}
\centering\small
\setlength{\tabcolsep}{15pt}
\fitwidth{%
\begin{tabular}{llrr}
\toprule
Axis & Configuration & Mean AUROC & $\Delta$ vs.\ published \\
\midrule
\textbf{Published (used throughout)} & last-token / residual stream / mean-difference & \textbf{0.9519} & --- \\
pooling & mean over all tokens & 0.8534 & $-0.099$ \\
pooling & mean over response tokens only & 0.9318 & $-0.020$ \\
pooling & max-pooling & 0.8847 & $-0.067$ \\
probe site & MLP output & 0.9514 & $-0.001$ \\
probe site & attention output & 0.9561 & $+0.004$ \\
centroid estimator & median & 0.9512 & $-0.001$ \\
centroid estimator & 20\%-trimmed mean & 0.9512 & $-0.001$ \\
\bottomrule
\end{tabular}
}
\end{table*}

\FloatBarrier
\section{Sample Efficiency}
\label{app:sampleeff}

\paragraph{Generalisation holds across base-model scale.} Re-running the calibrate-once leave-one-distribution-out test (the paper's Generalisation Beyond the Calibration Distribution section, the paper's calibrate-once leave-one-distribution-out table) with the pooled direction fit on Llama-3.2-3B and Llama-3.1-8B gives LODO mean AUROC \rcpt{0.761} and \rcpt{0.758}, versus \rcpt{0.728} at 1B, a small $1\text{B}\to3\text{B}$ lift that then plateaus. The in-domain per-distribution ceiling rises monotonically (\rcpt{0.849}$\to$\rcpt{0.859}$\to$\rcpt{0.869}) but the leave-one-out gap does not shrink, so calibrate-once transfer is a property of the pooled direction, not of base-model capacity. \textsc{trident}-law remains the weak spot at every scale.

\begin{figure}[htbp]
\centering
\includegraphics[width=0.62\columnwidth]{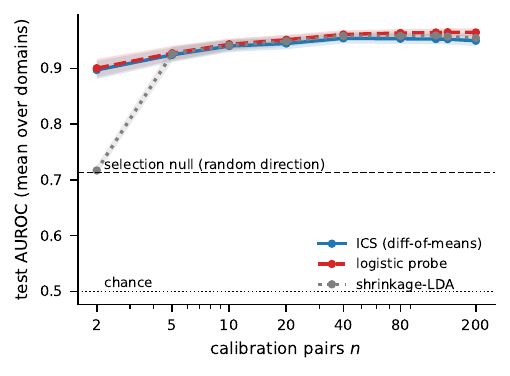}
\caption{Sample efficiency (mean over 20 ablated domains, $\pm$1 s.e.). The logistic probe ties or beats ICS at every calibration size, there is no small-$n$ crossover, and shrinkage-LDA is catastrophic at $n{=}2$. All estimators are read against the selection null (0.71), not chance.}
\label{fig:sampleeff}
\end{figure}

The hypothesised small-$n$ advantage of the closed-form estimator does not exist. A regularised logistic probe ties or beats ICS at every calibration size $n \in \{2,\ldots,200\}$ (Figure~\ref{fig:sampleeff}; at $n{=}10$: 15/20 domains, mean $+0.003$); shrinkage-LDA and whitening buy nothing (significant in 0--1/20) and shrinkage is catastrophic at $n{=}2$. What survives is a fitting-cost claim, not an accuracy claim: ICS is approximately the small-$n$ solution of the regularised probe, obtained in closed form with no gradient step and no regularisation sweep. It is not free of labelled supervision beyond the calibration pairs; layer and threshold selection each consume a labelled validation slice, exactly as they do for the logistic probe. CCS, run zero-supervision, reaches 0.917 with labelled layer selection, collapses to 0.69 under its own unsupervised layer choice.

\FloatBarrier
\section{Further Scope Conditions}
\label{app:pitfalls}
Four additional bounds, each stated briefly in the paper's Scope and Limitations section and given in full here. \textbf{First-party access} is a cost advantage for self-monitoring, not third-party auditing, and \emph{performed} compliance is not evaluated adversarially (\citealp{panfilov2025strategic,needham2025evalaware}, the paper's Comparison with Deployed Guards section). \textbf{The degeneracy diagnosis} is specific to compliance benchmarks, since the safety suite is cleaner at 2/7 degenerate (\S\ref{sec:degeneracy}). \textbf{ICS needs the scored text} to end on the case narrative (\S\ref{sec:calib}). \textbf{Absolute probe numbers} sit on \rcpt{0.71}--\rcpt{0.77} selection floors, not $0.5$.

\FloatBarrier
\section{Extended Related Work}
\label{sec:discussion-related}
\label{subsec:related-work}
The full positioning against prior work, condensed to a few sentences in the paper's Introduction section. The independent report of the same rule-blindness pattern for safety-relevant activation probes (the paper's Introduction section) extends to two further concurrent findings, Legilimens \citep{deng2024legilimens} and Gavel \citep{gavel2026}, on top of the two already named there.

Difference-of-means and related linear readouts underpin representation engineering \citep{zou2023representation}, the refusal direction \citep{arditi2024refusal}, and mass-mean probing \citep{marks2023geometry}, with contrast-consistent search \citep{burns2023discovering} offering an unsupervised variant, and safety properties have been shown to occupy multi-dimensional rather than single directions \citep{zhao2025harmfulness,pan2025hidden}. We apply the same estimator as a monitor and ask what it reads when classes are defined by written rules rather than safety labels. Recent systems read policy adherence from hidden states \citep{rachmil2025whitening,siren2026}, including DataShield's training-sample filter \citep{datashield2026}, whose inverted valence motivates our term \emph{adherence}. A parallel line ships \emph{dynamic}, policy-conditioned guards that re-target to a new policy at inference without retraining \citep{hoover2025dynaguard}, and our \emph{frozen} characterisation (the paper's Generalisation Beyond the Calibration Distribution section) applies only to the static classifiers benchmarked here, leaving open whether such dynamic systems pass our counterfactual rule test. We benchmark ICS against Llama Guard~3, WildGuard, Qwen3Guard, HarmBench \citep{mazeika2024harmbench}, SIREN \citep{siren2026}, GLiGuard, and Latent Policy Guard \citep{li2026lpg}, under lexical floors, a selection null, and benign inputs, controls none of them report, extending them to the never-pooled and degeneracy-audit benchmarks of \S\ref{sec:guardgap}. A training-free gradient signal \citep{xie2024gradsafe} and a concurrent activation probe \citep{schwarz2026entanglement} draw the same efficiency argument, the latter independently reaching our shared-blind-spot conclusion for safety harm, that activation scores read a broad risk register rather than the rule-context relation, a result we establish here for \emph{regulatory compliance} and extend with a difference-of-means alternative, a cross-family result (the paper's Generalisation Beyond the Calibration Distribution section), and a selection deployment (the paper's Score-Guided Candidate Selection section). Among these, Rachmil et al.\ \citep{rachmil2025whitening} are the nearest neighbour, compliance-framed like ours, warranting direct contrast. The methods differ in estimator, our difference of means against their whitened out-of-distribution score, and in readout, our last-token residual against their full-context representation. On their own benchmark our truncation-limited last-token probe lands at chance, AUROC 0.46 with an oracle best layer topping 0.53, since typical cases exceed our context window while their full-context method does not, a result we detail together with a lexical-shortcut audit of that benchmark in \S\ref{sec:degeneracy}. The methods are complementary, ICS a zero-cost read for short cases and whitening the appropriate tool for long transcripts, and our rule-blindness finding (the paper's What the Score Measures section) extends to their detector class too.

Whether such probes and the benchmarks used to evaluate them can be trusted at all is a second, related question. Probes degrade under shift \citep{kumar2026pressure}, pair selection dominates gains \citep{natarajan2026pairs}, models evade probes \citep{mcguinness2025chameleons}, and obfuscated activations defeat monitors \citep{bailey2024obfuscated}, while linear probes more broadly lean on surface evidence rather than the property they claim to read \citep{boxo2025linearprobes}. Linear probes for strategic deception report AUROC 0.96 to 0.999 with neither a lexical floor nor a random-direction control \citep{goldowskydill2025deception}, the identical un-floored pattern in a second subfield. The lexical-shortcut check is established practice, not a contribution here. TF-IDF baselines expose construction artifacts in adjacent jailbreak and hallucination benchmarks \citep{hussain2026parallax}, and are read elsewhere as good news for cheap CPU guardrails rather than as a floor \citep{guardchain2025}, a sign we flip explicitly. Our \emph{rule-only} and \emph{no-rule} conditions are premise-only and hypothesis-only baselines \citep{poliak2018hypothesis}, and the degeneracy we document is the shortcut-learning failure mode \citep{mccoy2019hans} recurring in rule-conditioned compliance, established here for guards as well as probes, and motivating the counterfactual rule-following tests \citep{sun2024rulebench} current benchmarks lack. Control tasks \citep{hewitt2019control} audit probe \emph{capacity}, and our budget-matched nulls extend that logic to the hyperparameter search. Our corpora include OmniCompliance \citep{hu2026omnicompliance}, AIReg-Bench \citep{aireg2025}, CompliBench \citep{yang2026complibench}, and FlexBench \citep{ding2026flexguard}, of which four of seven benchmarks tested cannot support the measurement at all. A convergent audit \citep{land2026auditing} asks whether benchmark \emph{labels} are correct from item-response patterns across models, and we ask instead whether correct labels measure the construct claimed, from the item text alone.

\bibliographystyle{unsrt}
\bibliography{references}

\end{document}